%% file: main.tex
\documentclass{article}

\usepackage{microtype}
\usepackage{graphicx}
\usepackage{subcaption}
\usepackage{booktabs} 

\usepackage{hyperref}

\usepackage[preprint]{icml2026}

\usepackage{amsmath}
\usepackage{amssymb}
\usepackage{mathtools}
\usepackage{amsthm}

\usepackage[capitalize,noabbrev]{cleveref}

\theoremstyle{plain}

\theoremstyle{definition}

\theoremstyle{remark}

\usepackage[textsize=tiny]{todonotes}

\usepackage{booktabs}   
\usepackage{multirow}
\usepackage{siunitx}    

\newcommand{\best}[1]{\textbf{#1}}

\icmltitlerunning{Is Retraining-Free Enough? The Necessity of Router Calibration for Efficient MoE Compression}

\begin{document}

\twocolumn[
  \icmltitle{Is Retraining-Free Enough? The Necessity of Router Calibration \\for Efficient MoE Compression}



  \icmlsetsymbol{equal}{*}

  \begin{icmlauthorlist}
    \icmlauthor{Sieun Hyeon}{yyy}
    \icmlauthor{Jaeyoung Do}{yyy,sch}


  \end{icmlauthorlist}

  \icmlaffiliation{yyy}{Department of Electrical and Computer Engineering, Seoul National University, Seoul, South Korea}
  \icmlaffiliation{sch}{Interdisciplinary Program in Artificial Intelligence, Seoul National University, Seoul, South Korea}

  \icmlcorrespondingauthor{Jaeyoung Do}{jaeyoung.do@snu.ac.kr}


  \vskip 0.3in
]

\printAffiliationsAndNotice{}  

\begin{abstract}
    Mixture-of-Experts (MoE) models scale capacity efficiently, but their massive parameter footprint creates a deployment-time memory bottleneck. We organize retraining-free MoE compression into three paradigms—Expert Pruning, Expert Editing, and Expert Merging—and show that persistent post-compression degradation largely stems from a neglected factor: router–expert mismatch when experts are changed but the router is left untouched. We argue that effective \textit{retraining-free} compression should avoid updating expert parameters while allowing lightweight router calibration. To this end, we propose Router Knowledge Distillation (Router KD), which updates only a tiny fraction of parameters (the router) by distilling the original model’s next-token distribution on unlabeled calibration data. Experiments across representative methods in all three paradigms demonstrate consistent performance recovery, with substantially larger gains in fine-grained MoEs (many small experts) than in coarse-grained MoEs due to their more complex routing decision boundaries.

\end{abstract}

\begin{figure*}[t]
    \centering 
    \includegraphics[width=\linewidth]{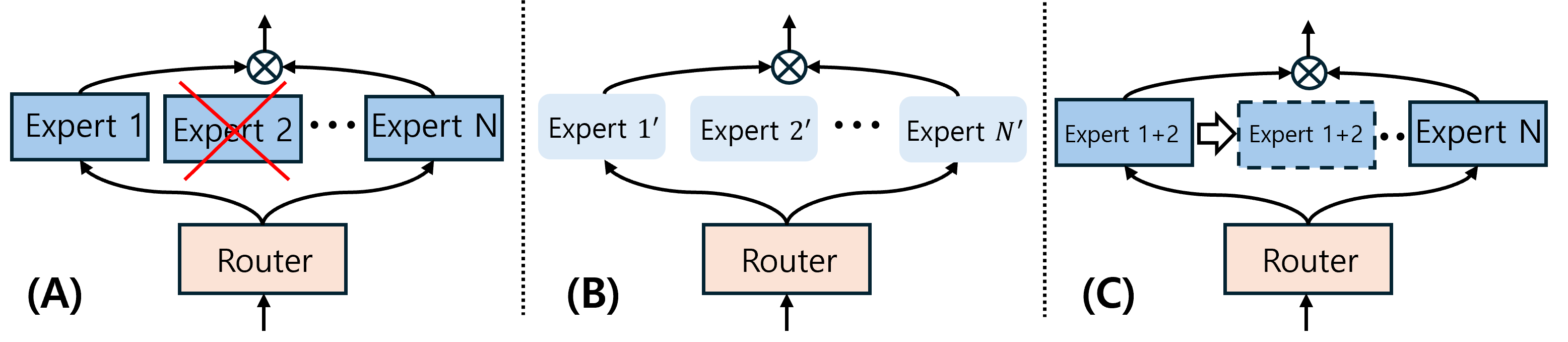}
    \caption{Illustration of the three MoE compression paradigms. The diagram depicts (A) \textit{Expert Pruning} (selection), (B) \textit{Expert Editing} (decomposition or modification), and (C) \textit{Expert Merging} (aggregation) to reduce model size.}
    \label{fig:main_fig} 
\end{figure*}

\section{Introduction}

Large Language Models (LLMs) have driven a paradigm shift in artificial intelligence, exhibiting remarkable capabilities across a broad spectrum of tasks from creative generation and code synthesis to complex mathematical reasoning. \cite{gpt5} Beyond natural language processing, their impact now extends to domains such as robotics, medicine, and scientific discovery. Motivated by empirical scaling laws \cite{scalinglaws} that link increased model capacity to improved performance, recent progress has been dominated by ever-larger models. However, the rapid growth in parameter counts has exposed fundamental challenges in efficiency, memory footprint, and deployability.

The Mixture-of-Experts (MoE) \cite{MoE1, MoE3} architecture has emerged as a key response to this tension. By decoupling total model capacity from per-token computation, MoE enables models to scale to massive parameter counts while activating only a sparse subset of experts at inference time. This property has made MoE a cornerstone of modern foundation models, offering an appealing balance between performance and computational efficiency \cite{mixtral, qwen3technicalreport}. Despite these advantages, MoE models introduce a severe memory bottleneck: the full parameter set must still be resident in memory, even though only a fraction is used per token \cite{MoE2}. As a result, MoE LLMs remain prohibitively expensive to deploy in resource-constrained environments, limiting their accessibility to most practitioners and users.

To mitigate this issue, a growing body of work has focused on \emph{retraining-free} compression of MoE architectures, methods that reduce memory consumption without full-scale retraining \cite{reap, hcsmoe, td_moe}. Since experts account for the overwhelming majority of parameters in MoE LLMs, existing approaches have primarily targeted expert-side compression. These methods are often presented as mutually competing solutions that optimize compression ratio while minimizing performance degradation, and the field has rapidly become crowded with claims of near-optimal efficiency.

However, despite this surge of innovation, a fundamental question remains unresolved: why does performance degradation persist even when expert compression is carefully designed? We argue that the core limitation is not the absence of a ``perfect'' expert compression method, but rather a systematic mismatch between compressed experts and an unmodified router. Expert compression, whether by removal, modification, or merging, inevitably perturbs the functional landscape on which routing decisions were originally learned. Yet, in almost all retraining-free approaches, the router is left unchanged. This mismatch leads to suboptimal expert selection and amplifies performance loss.

In this work, we formalize this observation by systematizing retraining-free MoE compression methods into three categories: \emph{Expert Pruning}, \emph{Expert Editing}, and \emph{Expert Merging}. Using this taxonomy, we investigate a largely overlooked dimension of MoE compression: \emph{router calibration}. Through theoretical analysis, we show that routing discrepancies arise even in best-case compression scenarios, and that these discrepancies compound across layers. Our empirical analysis further demonstrates that router miscalibration is a dominant contributor to post-compression performance degradation across all three compression paradigms.

These findings lead to a critical re-examination of the notion of ``retraining-free'' compression. We show that fully retraining-free compression---defined as leaving both experts and router untouched---is often impractical for achieving strong performance. Instead, we advocate a more precise interpretation: avoiding updates to expert parameters, while allowing lightweight router recalibration.

To this end, we introduce \emph{Router Knowledge Distillation (Router KD)} as a simple yet effective recovery mechanism. Router KD updates only the router parameters of the compressed model, distilling knowledge from the original model's output distribution. Importantly, this process incurs minimal computational overhead, as the router constitutes only a tiny fraction of the total model parameters. We apply Router KD to representative methods from each compression category and quantify how much performance can be recovered by calibrating the router alone.

Our experimental results show that Router KD consistently and substantially mitigates performance degradation across Expert Pruning, Editing, and Merging. Moreover, we reveal that the effectiveness of router calibration depends strongly on MoE architecture. In particular, fine-grained MoE models with many small experts (e.g., \textit{Qwen3-30B-A3B-Instruct} \cite{qwen3technicalreport}) benefit significantly more from Router KD than coarse-grained models with fewer, larger experts (e.g., \textit{Mixtral-8$\times$7B-Instruct} \cite{mixtral}), due to the increased complexity and flexibility of their routing decision boundaries. Our contributions are summarized as follows:

\begin{itemize}
    \item We propose a taxonomy of retraining-free MoE compression---Expert Pruning, Editing, and Merging---and identify router miscalibration as a primary source of post-compression performance degradation. Through theoretical and empirical analysis, we show that expert compression must be coupled with router calibration for effective performance preservation.
    \item We introduce Router Knowledge Distillation as a lightweight and general recovery strategy that updates only router parameters. Extensive experiments demonstrate that Router KD consistently restores performance across all compression paradigms and is particularly effective for fine-grained MoE architectures.
\end{itemize}

\section{Related Works}

We categorize retraining-free MoE compression into \textit{Expert Pruning}, \textit{Editing}, and \textit{Merging}. Crucially, we focus exclusively on \textit{parameter-level} compression—reducing the number of parameters—and explicitly exclude bit-level compression techniques such as quantization. \textbf{Expert Pruning} removes redundant experts ($k$). Strategies include minimizing reconstruction loss~\cite{NAEE}, differentiable selection~\cite{bai2025diep}, and leveraging router magnitudes~\cite{reap}. Other approaches utilize output discrepancy bounds~\cite{anonymous2025compressing}, token variation~\cite{easy_ep}, trajectory-based importance~\cite{moepathfinder}, or coarser layer-level pruning~\cite{mc_suite, RS_TMLR}. \textbf{Expert Editing} compresses expert internals via decomposition while retaining the expert count. Methods employ Singular Value Decomposition (SVD)~\cite{moe-svd}, factorization into shared and specific components~\cite{molae}, rank decomposition with shared bases~\cite{mobe}, or tensor decomposition~\cite{td_moe}. \textbf{Expert Merging} is grounded in model merging hypotheses~\cite{modelsoup, mergingmodelsfisher}. This approach synthesizes experts via output similarity clustering~\cite{hcsmoe}, selective dual-masks~\cite{puzzlemoe}, compression matrices~\cite{mergemoe}, or importance-guided coefficient merging~\cite{expertmerging}. See Appendix~\ref{appendix:appendix} for extended related works.

\section{Causes of Performance Degradation}
\label{sec:degradation}

Despite extensive research on MoE compression, a critical unresolved issue is the performance degradation observed in compressed models compared to their original counterparts. While this might be perceived as an inevitable trade-off due to the reduction in parameters, minimizing such performance loss remains the primary challenge in MoE compression. 
In this section, we isolate a key contributor: \emph{router--expert mismatch}. We derive how compression perturbs routing behavior and show that router-induced error can arise even under favorable compression scenarios, compounding across layers.

\subsection{Original MoE LLMs}

In the original MoE model before compression, an input ($ x $) first passes through the gate network, producing $n$ \textit{expert activation scores}. Depending on the implementation, the gate network outputs either the raw logits or the probabilities after a softmax operation. Let these $n$ \textit{expert activation scores} be denoted as ($ g_0, \ldots, g_n $). Likewise, let the $n$ experts in the same layer as the gate network be represented as ($ E_0, \ldots, E_n $). Assume that this MoE model activates the top-$k$ experts. In this case, the computation for the input $x$ is performed as follows.

$$
\mathcal{S} \subset \{0, 1, \ldots, n-1\}, \quad |S| = k, \quad k < n
$$

We define the renormalized expert activation weights as
$$
\tilde{g}_i = \frac{g_i}{\sum_{j \in \mathcal{S}} g_j}, \quad i \in \mathcal{S}
$$

Using these normalized weights, the output of the MoE layer for input ($x$) is computed as:
$$
y = \sum_{i \in \mathcal{S}} \tilde{g}_i \cdot E_i(x).
$$

\subsection{Expert Pruning ($N \rightarrow N-\alpha$)}

When the original MoE model undergoes pruning, the inference can fall into one of the following three scenarios:

1. Best scenario: All experts selected by the original model remain available after pruning, and the same experts are used without any change.

2. Most common scenario: Pruning is imperfect, and while some originally selected experts remain available and are used as before, others are removed. For the dropped experts, the model must instead rely on alternative experts as substitutes.

3. Worst scenario: All experts that were originally selected are pruned out, forcing the model to replace every originally chosen expert with different ones.

Let

\begin{itemize}
    \item $\mathcal{S}$ be the set of expert indices selected by the original MoE model,
    \item $\mathcal{P} \subseteq \{0, \ldots, n-1\}$ be the set of experts that remain after pruning, and $|\mathcal{S}| \leq |\mathcal{P}|$.
    \item $\mathcal{S}'$ be the set of experts effectively used by the pruned model.
\end{itemize}
We analyze the relationship between $\mathcal{S}$, $\mathcal{P}$ and $\mathcal{S}'$.

\noindent\textbf{Gate scores before and after pruning.}
We denote by $g_i^{\mathrm{orig}}(x)$ the expert activation scores produced by the original gate network for an input $x$, and by $g_i^{\mathrm{pruned}}(x)$ the scores produced when running the pruned model end-to-end.
For any index set $\mathcal{A}$, we define the corresponding renormalized weights as
$$
\tilde{g}_i^{\mathrm{orig},(\mathcal{A})}(x)
= \frac{g_i^{\mathrm{orig}}(x)}{\sum_{j \in \mathcal{A}} g_j^{\mathrm{orig}}(x)} 
$$

$$
\tilde{g}_i^{\mathrm{pruned},(\mathcal{A})}(x)
= \frac{g_i^{\mathrm{pruned}}(x)}{\sum_{j \in \mathcal{A}} g_j^{\mathrm{pruned}}(x)},
\quad i \in \mathcal{A}.
$$

\subsubsection{Best Scenario}

In the best scenario, all originally selected experts remain after pruning:

$
\mathcal{S} \subseteq \mathcal{P}
\quad\Longrightarrow\quad
\mathcal{S}' = \mathcal{S}.
$

The original and pruned MoE outputs for input $x$ can then be written as
$$
y_{\mathrm{orig}}(x)
= \sum_{i \in \mathcal{S}} \tilde{g}_i^{\mathrm{orig},(\mathcal{S})}(x)\, E_i(x)
$$

$$
y_{\mathrm{pruned}}^{\mathrm{best}}(x)
= \sum_{i \in \mathcal{S}} \tilde{g}_i^{\mathrm{pruned},(\mathcal{S})}(x)\, E_i(x).
$$

The difference between the original and pruned MoE outputs in the best scenario is then given by:

$$
\begin{aligned}
& \left\|y_{\mathrm{orig}}(x) - y_{\mathrm{pruned}}^{\mathrm{best}}(x) \right\|=\\
& \left\|\sum_{i \in \mathcal{S}} \tilde{g}_i^{\mathrm{orig},(\mathcal{S})}(x)\, E_i(x) - \sum_{i \in \mathcal{S}} \tilde{g}_i^{\mathrm{pruned},(\mathcal{S})}(x)\, E_i(x)\right\|=\\
& \left\| \sum_{i \in \mathcal{S}} \left( \tilde{g}_i^{\mathrm{orig},(\mathcal{S})}(x) - \tilde{g}_i^{\mathrm{pruned},(\mathcal{S})}(x) \right) E_i(x) \right\|
\end{aligned}
$$

In this scenario, the set of active experts is identical to that of the original model. However, note that even if the selected experts match exactly pre- and post-pruning, the router's output values are unlikely to be identical. Because MoE LLMs are multi-layered, for the router outputs to be perfectly identical, this best-case scenario must be satisfied in every layer, which is statistically improbable. Therefore, it can be trivially stated that $\tilde{g}_i^{\mathrm{orig},(\mathcal{A})}(x) \neq \tilde{g}_i^{\mathrm{pruned},(\mathcal{A})}(x)$

\subsubsection{Most Common \& Worst Scenario}

For the mathematical formulation of the Most Common and Worst scenarios, please refer to Appendix~\ref{appendix:equations_pruning}. A key insight from this analysis is that discrepancies induced by the router arise even in the rare best-case scenario. Although the router weight $\tilde{g}_i$ is merely a scalar value between 0 and 1, the Expert itself is a matrix consisting of over 1 million scalar values; thus, the resulting deviation is by no means negligible. Furthermore, this difference will inevitably amplify in the Most Common and Worst scenarios, where the selected experts differ from those in the original model. This suggests that the divergence between the pruned and original models is, in part, attributable to the router.

\subsection{Expert Editing ($N \rightarrow N$, Parameters $P \rightarrow P'$)}
Unlike pruning, Expert Editing preserves the total number of experts. However, as the parameters in preceding layers are modified, the router's computation results may shift. Consequently, for the same input, the edited model may select different experts compared to the original model. The inference process can fall into one of the following three scenarios:

1. Best scenario: The router's selection remains completely unchanged after editing, meaning the exact same experts chosen by the original model are utilized.

2. Most common scenario: Due to the router's altered expert activation scores following the editing process, the selection is partially changed. While some originally selected experts are retained and used as before, others are replaced by different experts.

3. Worst scenario: The router's output shifts drastically such that none of the originally selected experts are chosen, and a completely different set of experts is activated.

Let
\begin{itemize}
    \item $\mathcal{S}$ be the set of expert indices selected by the original MoE model,
    \item $\mathcal{S}^{\mathrm{edit}} \subseteq \{0, \ldots, n-1\}$ be the set of expert indices selected by the edited MoE model, and $|\mathcal{S}| = |\mathcal{S}^{\mathrm{edit}}|$.
    \item $E_i$ denote the original experts, and $X_i$ denote the corresponding edited experts obtained by modifying $E_i$.
\end{itemize}

\noindent\textbf{Gate scores before and after editing.}
As before, we denote by $g_i^{\mathrm{orig}}(x)$ the expert activation scores produced by the original gate network for an input $x$.
We denote by $g_i^{\mathrm{edit}}(x)$ the activation scores produced when running the edited MoE model end-to-end on the same input.
For any index set $\mathcal{A}$, we define the renormalized weights for the edited model as
$$
\tilde{g}_i^{\mathrm{edit},(\mathcal{A})}(x)
= \frac{g_i^{\mathrm{edit}}(x)}{\sum_{j \in \mathcal{A}} g_j^{\mathrm{edit}}(x)},
\quad i \in \mathcal{A}.
$$

\subsubsection{Best Scenario}

In the best scenario, the router's top-$k$ selection remains unchanged even after expert editing, i.e.,
$$
\mathcal{S}^{\mathrm{edit}} = \mathcal{S}.
$$
The original MoE output for input $x$ is
$$
y_{\mathrm{orig}}(x)
= \sum_{i \in \mathcal{S}} \tilde{g}_i^{\mathrm{orig},(\mathcal{S})}(x)\, E_i(x),
$$
while the edited MoE output in the best scenario becomes
$$
y_{\mathrm{edit}}^{\mathrm{best}}(x)
= \sum_{i \in \mathcal{S}} \tilde{g}_i^{\mathrm{edit},(\mathcal{S})}(x)\, X_i(x).
$$
Even though the index set of selected experts is identical, both the gate scores and the expert functions may differ between $y_{\mathrm{orig}}(x)$ and $y_{\mathrm{edit}}^{\mathrm{best}}(x)$.

The difference between the original and edited MoE outputs in the best scenario is then given by:

$$
\begin{aligned}
& \left\|y_{\mathrm{orig}}(x) - y_{\mathrm{edit}}^{\mathrm{best}}(x) \right\|=\\
& \left\|\sum_{i \in \mathcal{S}} \tilde{g}_i^{\mathrm{orig},(\mathcal{S})}(x)\, E_i(x) - \sum_{i \in \mathcal{S}} \tilde{g}_i^{\mathrm{edit},(\mathcal{S})}(x)\, X_i(x)\right\| \\
& \left\|\sum_{i \in \mathcal{S}} \left( \tilde{g}_i^{\mathrm{orig},(\mathcal{S})}(x)\, E_i(x) - \tilde{g}_i^{\mathrm{edit},(\mathcal{S})}(x)\, X_i(x)\right)\right\|
\end{aligned}
$$

Even if sophisticated mathematical approximation techniques render $E_i$ and $X_i$ nearly identical, they are not strictly identical; thus, $g_i^{\mathrm{orig}}$ and $g_i^{\mathrm{edit}}$ cannot be equal. Given that each expert is a matrix containing over 1 million parameters, even slight deviations in router outputs—despite the similarity between $E_i$ and $X_i$—inevitably lead to non-negligible differences in the final output.

\subsubsection{Most Common \& Worst Scenario}

For the mathematical formulation of the Most Common and Worst scenarios, please refer to Appendix~\ref{appendix:equations_editing}. These equations imply that the greater the imperfection in the Expert Editing method, the more the router's output diverges from the original, inevitably altering the set of selected experts. Furthermore, it can be inferred that the error stemming from selecting an incorrect expert—specifically the difference between the original expert $E_i$ and a mismatched edited expert $X_j$—exceeds the error introduced by the editing process itself (i.e., the difference between $E_i$ and its corresponding $X_i$). Given the inherent nature of Sparse MoE, where experts within the same layer exhibit significant distinctiveness, the divergence between completely different experts (i.e., $E_i$ and $X_j$) is highly likely to surpass the deviation between an original expert and its approximated edited counterpart (i.e., $E_i$ and $X_i$). Consequently, if router calibration can mitigate the risk of selecting $X_j$ and ensure the selection of $X_i$, the deviation from the original output (relative to $E_i$) can be minimized.

\subsection{Expert Merging ($N \rightarrow M$, where $M < N$)}

Unlike Pruning and Editing (three scenarios each), Expert Merging entails nine distinct scenarios. Due to space limitations, we refer the reader to Appendix~\ref{appendix:equations_merging} for the detailed mathematical formulations. To summarize, similar to Pruning and Editing, Expert Merging inevitably suffers from router-induced errors even in the best-case scenario. Furthermore, in the most common scenario, the merged experts may be selected differently than originally intended by the router, potentially resulting in even greater discrepancies.

\begin{figure}[t]
    \centering
    \begin{subfigure}[b]{0.48\textwidth}
        \centering
        \includegraphics[width=\linewidth]{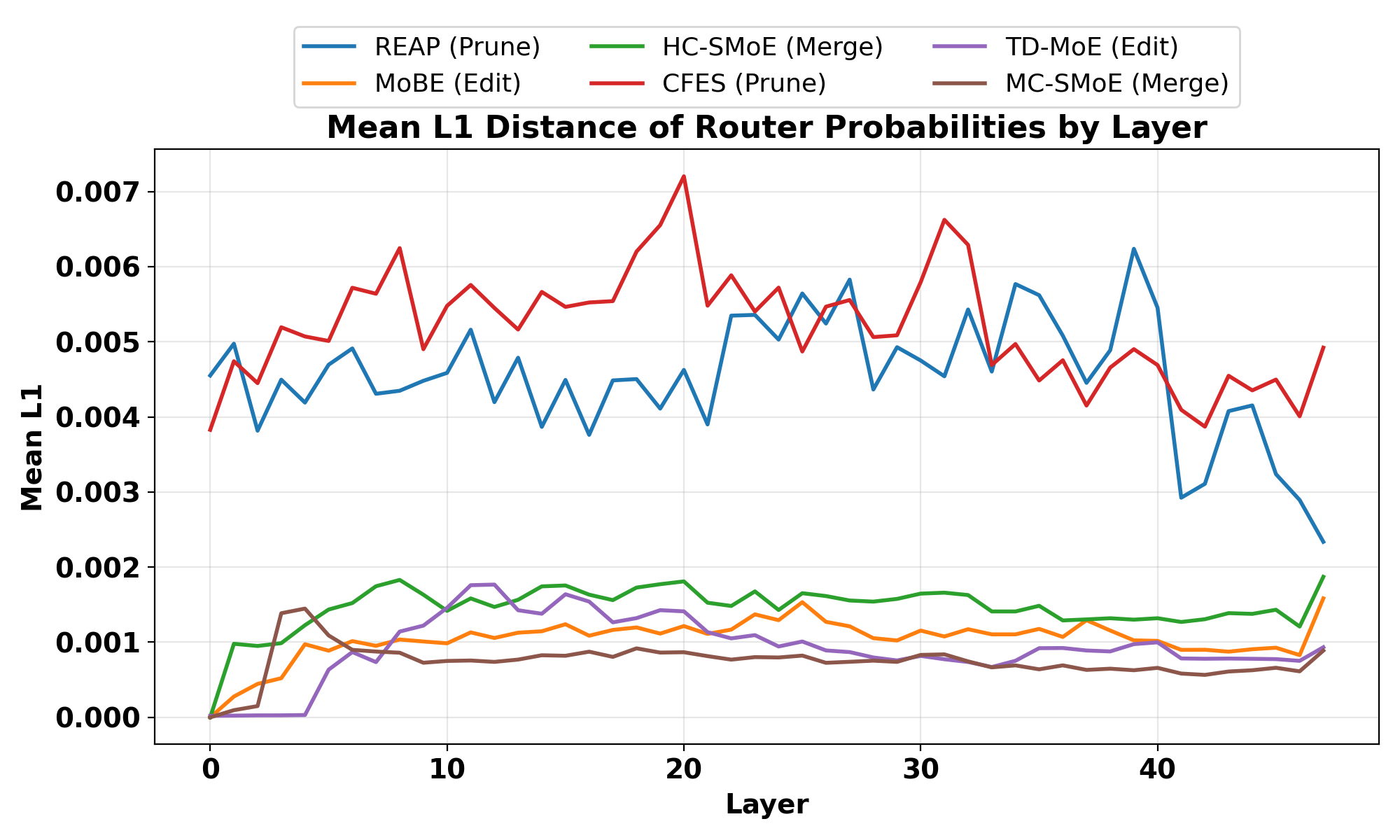}
        \caption{Mean L1 Distance of Routing Probabilities}
        \label{fig:L1_Distance}
    \end{subfigure}
    \hfill
    \begin{subfigure}[b]{0.48\textwidth}
        \centering
        \includegraphics[width=\linewidth]{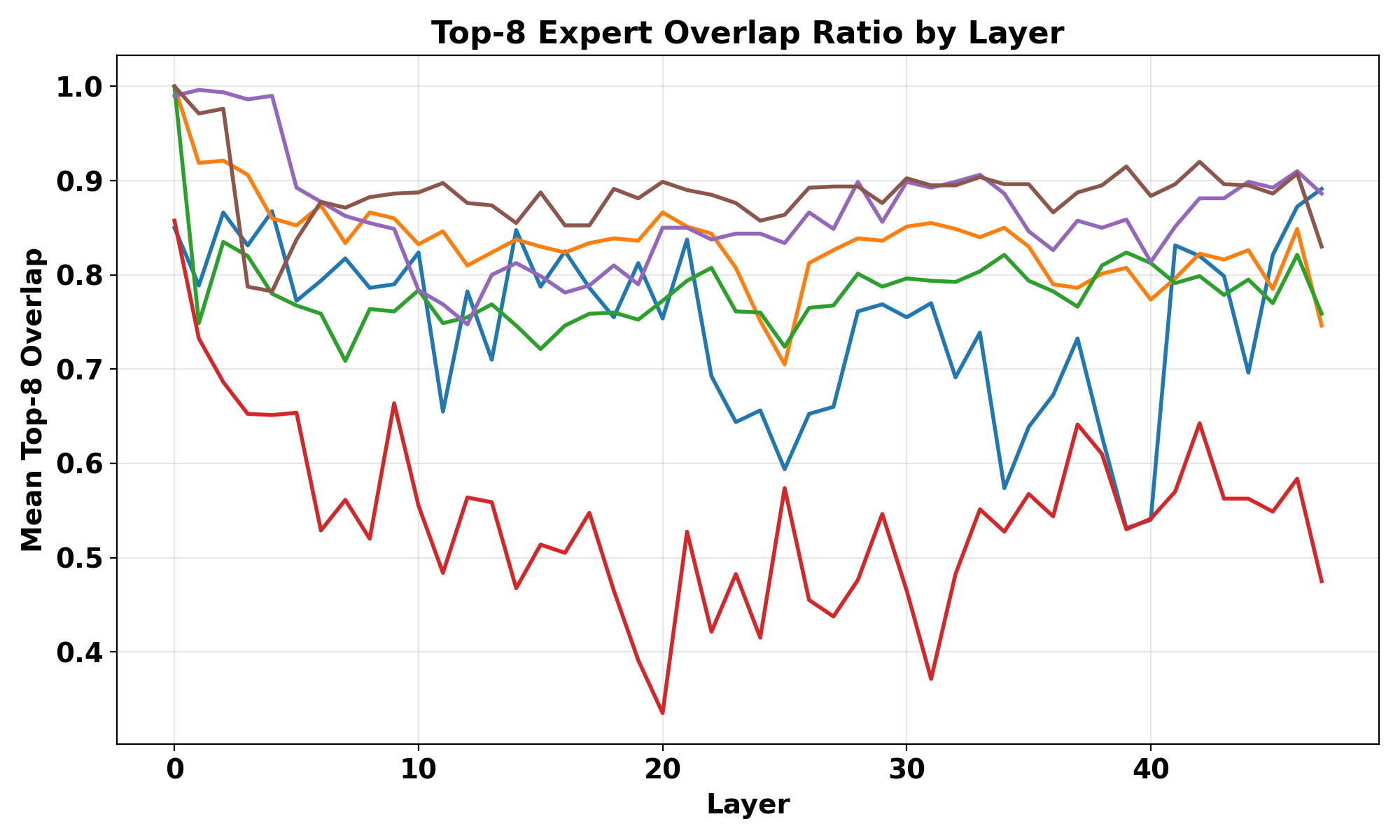}
        \caption{Top-8 Expert Overlap Ratio}
        \label{fig:Top_8_Expert_Overlap}
    \end{subfigure}
    
    \caption{Layer-wise Analysis of Router Behavior on \textit{Qwen3-30B-A3B-Instruct-2507}(128 Experts). Results are based on 100 samples from the ELI5 dataset. 
    (a) shows the L1 distance of routing probabilities, where Pruning methods exhibit larger divergence compared to Editing and Merging. 
    (b) illustrates the Top-8 expert overlap ratio, indicating that router consistency degrades in deeper layers across all compression methods.}
    \label{fig:Router_Analysis}
\end{figure}

\subsection{Empirical Analysis of Router Behavior}

We empirically examine how routing changes after compression. Using \textit{Qwen3-30B-A3B-Instruct-2507} as the backbone, we sample 100 questions from ELI5~\cite{eli5} and compare the router outputs between the original and compressed models.

As shown in Figure~\ref{fig:L1_Distance}, the routing probabilities assigned to experts deviate from the original model across most layers for all three methods: Pruning, Editing, and Merging. Notably, Pruning models exhibit a relatively larger divergence in assigned probabilities compared to Editing and Merging. This is attributed to the fact that Pruning drops experts and masks the corresponding indices, whereas Editing and Merging retain the total number of experts, thereby preserving the dimensionality of the router.

The impact of the router post-compression is even more pronounced in Figure~\ref{fig:Top_8_Expert_Overlap}, which illustrates the overlap ratio of experts selected by the router before and after compression. We observe that as the layers deepen, the router increasingly selects experts different from those intended by the original model. 
Together, these results support our theoretical analysis: preserving compressed MoE performance requires not only expert-side compression but also \emph{router calibration} to mitigate mismatch.

\section{Router Knowledge Distillation (Router KD)}

Motivated by our analysis in Section~\ref{sec:degradation}, we seek to {recalibrate} the routing behavior of a compressed MoE model so that it better reproduces the original model's next-token predictions {under fixed compressed experts}. To this end, we propose {Router Knowledge Distillation (Router KD)}, a lightweight distillation procedure that updates {only} the student router parameters while keeping all other student parameters frozen.

\paragraph{Router-only KD objective.}
Let $\mathcal{D}_{\mathrm{cal}}$ be an unlabeled calibration corpus, and let
$\theta_T$ denote the parameters of the original ($T$eacher) MoE model.
The compressed ($S$tudent) MoE model has parameters $\theta_S=(\theta_R,\theta_E)$,
where $\theta_R$ denotes router (gating) parameters and $\theta_E$ denotes all remaining parameters
(including experts and shared blocks). We freeze $\theta_E$ and optimize only $\theta_R$:
\begin{equation}
\theta_R^\star \;=\; \arg\min_{\theta_R}\;
\mathbb{E}_{x \sim \mathcal{D}_{\mathrm{cal}}}\!\left[\mathcal{L}_{\mathrm{RKD}}(x;\theta_T,\theta_R)\right].
\label{eq:router_kd_obj}
\end{equation}

\paragraph{Distillation loss.}
For an input token sequence $x=(x_1,\dots,x_L)$, let $z_T^{(t)} \in \mathbb{R}^{|\mathcal{V}|}$ and
$z_S^{(t)} \in \mathbb{R}^{|\mathcal{V}|}$ denote the teacher and student vocabulary logits for predicting the next token
at position $t$ (i.e., conditioned on the prefix $x_{\le t}$).
With temperature $\tau>0$, we define the softened next-token distribution for a model
$M\in\{T,S\}$ as:
\begin{equation}
\label{eq:router_kd_soft_targets}
p_M^{(t)}(\cdot) \;=\; \mathrm{Softmax}\!\left(\frac{z_M^{(t)}}{\tau}\right).
\end{equation}

We then minimize the token-level KL divergence from teacher to student, masking padding tokens.
Let $m_{t+1}\in\{0,1\}$ be the loss mask for the target position $(t{+}1)$, and define the
normalizer $N_x = \sum_{t=1}^{L-1} m_{t+1} + \epsilon$, where $\epsilon$ is a small constant. The router KD loss for a sequence is:
\begin{equation}
\label{eq:router_kd_loss}
\mathcal{L}_{\mathrm{RKD}}(x;\theta_T,\theta_R)
=
\frac{\tau^2}{N_x}
\sum_{t=1}^{L-1} m_{t+1}\,
D_{\mathrm{KL}}\!\left(p_T^{(t)} \,\|\, p_S^{(t)}\right).
\end{equation}

\noindent
Importantly, although the distillation loss is defined on the output token distribution, gradients are backpropagated and applied \emph{exclusively} to the student router parameters $\theta_R$, while all expert and backbone parameters remain frozen. This design directly calibrates the routing behavior of the compressed model so that it better matches the teacher’s next-token predictions under fixed compressed experts. In practice, we follow standard MoE routing implementations in which gradients flow through the (soft) gating weights of the selected experts during the forward pass, but parameter updates are restricted to the router.

By distilling output logits rather than matching router gate values explicitly, Router KD avoids requiring the teacher and student to share identical expert sets or gate dimensionalities, which may not hold after pruning, editing, or merging. Instead, the student router learns to route tokens to experts that best reproduce the teacher’s next-token distribution, thereby partially compensating for routing mismatch and functional discrepancies introduced by expert compression.

A key advantage of Router KD is its short wall-clock training time. Since only the router is updated, the number of trainable parameters is negligible compared to the full MoE model. Concretely, the router accounts for approximately $0.04\%$ of parameters in \textit{Qwen3-30B-A3B-Instruct-2507} and $0.002\%$ in \textit{Mixtral-8$\times$7B-Instruct-v0.1}. Accordingly, under the hyperparameter settings used in our experiments, Router KD required approximately 2 hours for Qwen3 and about 40 minutes for Mixtral, while yielding substantial performance recovery.

\section{Experiments}

\begin{figure*}[t]
    \centering 
    \includegraphics[width=\linewidth]{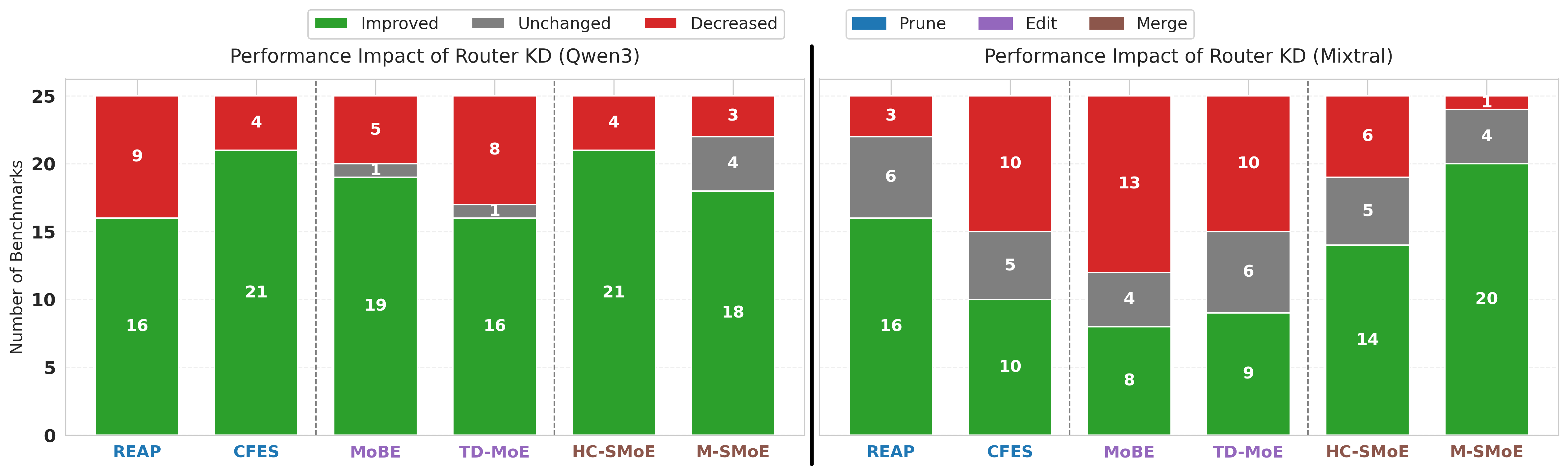}
    \caption{Performance Impact of Router KD on Qwen3 vs. Mixtral. The chart compares performance recovery across Pruning, Editing, and Merging, where green bars indicate improved benchmarks. Router KD proves significantly more effective for the fine-grained Qwen3 (Left) compared to the coarse-grained Mixtral (Right), which shows limited gains due to its simpler routing decision boundaries.}
    \label{fig:main_results} 
\end{figure*}

Our experiments are designed to answer the following questions:
\textbf{(i) Can Router KD consistently recover performance lost due to MoE compression?
(ii) Does its effectiveness generalize across different compression paradigms?
(iii) How does MoE architecture influence the benefits of router calibration?} To this end, we select two representative methods from each of the Expert Pruning, Editing, and Merging categories.
We evaluate Router KD on two widely used MoE backbones with contrasting architectures:
\textit{Qwen3-30B-A3B-Instruct-2507} and \textit{Mixtral-8$\times$7B-Instruct-v0.1}.
Hybrid compression methods are intentionally excluded to isolate the independent effect of router calibration.

\subsection{Baselines}

For \textbf{Expert Pruning} baselines, we adopt REAP~\cite{reap}, which defines expert importance based on a criterion that considers both router gate-values and the magnitude of expert outputs. We also compare against CFES~\cite{anonymous2025compressing}, which utilizes a coarse-to-fine expert selection strategy to efficiently identify critical experts by minimizing layer-wise output discrepancy. For \textbf{Expert Editing}, we utilize MoBE~\cite{mobe} as a baseline, which employs rank decomposition to separate experts into unique components and shared basis matrices to minimize reconstruction error. Although MoBE involves backpropagation and thus may not be strictly classified as retraining-free, we adopt it as a suitable baseline because it is significantly more computationally efficient than standard fine-tuning and does not require large-scale training datasets for compression. Additionally, we include TD-MoE~\cite{td_moe}, which treats experts as correlated tensors and applies Tucker Decomposition after aligning the data distribution via whitening. For \textbf{Expert Merging}, we select HC-SMoE~\cite{hcsmoe} as a baseline, which utilizes hierarchical clustering based on the similarity of expert outputs on calibration data. Furthermore, we employ M-SMoE~\cite{mcsmoe}, which integrates experts through activation frequency-based weighted averaging or permutation alignment to preserve collective knowledge.

\subsection{Benchmark Datasets}
We evaluated performance across various types of benchmark datasets to assess performance changes under diverse conditions. To assess general reasoning capabilities and QA, we used BBH \cite{bbh} (evaluated separately for both few-shot and zero-shot settings) and CoQA \cite{coqa}. For mathematical problem-solving and reasoning, we utilized GSM8k \cite{gsm8k}, GSM8k Platinum \cite{gsm8k_Platinum}, MATH \cite{hendrycksmath2021, lewkowycz2022solving, kydlicek2025fixing}, AIME 1983-2024 (averaged across all problems for those years) \cite{aime_1983_2024}, and AIME 2025 \cite{aime_2025}. Furthermore, we employed MBPP \cite{mbpp} and HumanEval-Instruct \cite{humaneval} to assess coding proficiency. In addition to these, we evaluated Chain-of-Thought (CoT) \cite{chainofthought} reasoning capabilities on the BBH, GSM8k, and GSM8k Platinum datasets (also evaluated separately for few-shot and zero-shot settings). Finally, nine benchmark datasets \cite{arc, hellaswag, medmcqa, medqa, OpenBookQA, piqa, WinoGrande, mmlu} were utilized to assess Multiple Choice Question Answering (MCQA) performance.

\subsection{Experimental Setup}
All experiments were evaluated using the lm-evaluation-harness \cite{eval-harness} and vLLM \cite{vLLM}. Additionally, Qwen3 was tested in an environment equipped with NVIDIA A100 40GB GPUs, while Mixtral was tested with A100 80GB GPUs. The random seed was fixed at 42 for all experiments, and all benchmarks were measured using greedy decoding with a temperature of 0. Hyperparameters for Router KD—including epochs, batch size, max length, number of calibration samples, and learning rate—were kept consistent across all experiments. A uniform expert retention rate of 62.5\% was applied; in the context of pruning, this entails a reduction in experts from 128 to 80 for Qwen3 and from 8 to 5 for Mixtral. For all baselines, Router KD used the same C4 \cite{c4} dataset, and the hyperparameters were set identically. Please refer to Appendix~\ref{appendix:expsetting} for more detailed information.

\subsection{Results}

We confirmed that Router KD effectively mitigates the performance degradation during model compression. Figure~\ref{fig:main_results} provides a visual summary of the experimental results, while detailed numerical data can be found in Appendix~\ref{appendix:exp_tables}. When applying Router KD to the Qwen3 backbone, we observed improvements across the majority of benchmarks. This trend was consistent across Expert Pruning, Editing, and Merging, demonstrating that using the teacher’s next-token distribution as supervision to optimize the student router is a sufficiently effective recovery strategy. However, this efficacy was not universal, and we identify specific scenarios where Router KD proved less effective.

\paragraph{Coarse-grained Experts}
The most significant disparity in the effectiveness of Router KD was observed when varying the backbone architecture. For Qwen3, Router KD improved benchmark scores across all six compression methodologies; conversely, the performance gains were relatively marginal for Mixtral. This distinction can be attributed to structural differences: Qwen3 (30.5B parameters) utilizes a \textit{fine-grained} expert structure with 128 experts per layer, whereas Mixtral (46.7B parameters) employs a \textit{coarse-grained} structure with only 8 experts per layer. In other words, while Qwen3 consists of many small experts, Mixtral comprises fewer, larger experts. Consequently, our results indicate that the beneficial impact of Router KD is diminished in coarse-grained MoE models like Mixtral.

Router KD can only refine the \emph{gating decision}---selecting a top-$k$ subset of experts and reweighting them to match the teacher.
Accordingly, its attainable gain is inherently limited when (i) the router has few alternative routing paths to switch to, and (ii) the teacher routing targets provide little additional information beyond an almost-hard decision.

\textit{(1) Small combinatorial routing space.}
Under top-$k$ routing with $E$ experts, the number of distinct expert subsets is $|\Omega| \;=\; \binom{E}{k}$. For Mixtral-style routing ($E{=}8$, $k{=}2$), $|\Omega|=\binom{8}{2}=28$, whereas for fine-grained MoEs (e.g., $E{=}128$, $k{=}8$), $|\Omega|=\binom{128}{8}\approx 1.43\times 10^{12}$. Hence, even if KD improves the router, the number of qualitatively different routing paths is fundamentally bounded in small-$E$ MoEs, since the discrete choice of the top-$k$ subset is limited even though reweighting within a subset is continuous.

\textit{(2) Gradient viewpoint: limited degrees of freedom and weaker KD signal.}
For intuition, consider the temperature-scaled \emph{routing} distributions over experts
$g_T(x), g_S(x)$, defined as $g_M(x) \;=\; \mathrm{Softmax}\!\left(z_M(x)/\tau\right)$, $z_M(x)\in\mathbb{R}^{E}$, $M\in\{T,S\}$ where $z_M(x)$ are router logits.
For the KL objective $\mathcal{L}=\mathrm{KL}(g_T\|g_S)$, the gradient w.r.t.\ each student logit is approximately
\begin{equation}
\frac{\partial}{\partial z_{S,e}}\,\mathrm{KL}\!\left(g_T \,\|\, g_S\right)
\;\approx\;
\frac{1}{\tau}\Big(g_{S,e}-g_{T,e}\Big)
\end{equation}
so the learning signal decomposes across the expert dimension $e\in\{1,\dots,E\}$.
If the router is linear, $z_S(x)=W_S x$, then
\begin{equation}
\nabla_{W_S}\mathcal{L}
\;\propto\;
\Big(g_{S,e}-g_{T,e}\Big)\,x^\top
\end{equation}
highlighting that the router receives adjustment opportunities along $E$ expert coordinates.

In coarse-grained MoEs, experts typically have broad coverage, and the teacher routing distribution is often highly concentrated (low entropy),
which means $H(g_T^{\text{coarse}}) \ll H(g_T^{\text{fine}})$.
When $g_T$ is already near-hard, $(g_S-g_T)$ quickly becomes small once the student matches the dominant experts, which deprives the gradients of informative `dark knowledge' regarding non-selected experts.
Moreover, when $E$ is small, the router has fewer coordinates to reshape and fewer alternative top-$k$ subsets to switch to, further limiting the observable benefit of router-only KD.
By contrast, in fine-grained MoEs with large $E$ and $E\!\gg\!k$, the routing space is vastly larger and the teacher targets are often less peaky, so KD provides richer `dark knowledge' over many non-selected experts, acting as a navigational cue across a substantially more complex decision boundary.

\begin{figure}[t]
    \centering 
    \includegraphics[width=0.5\textwidth]{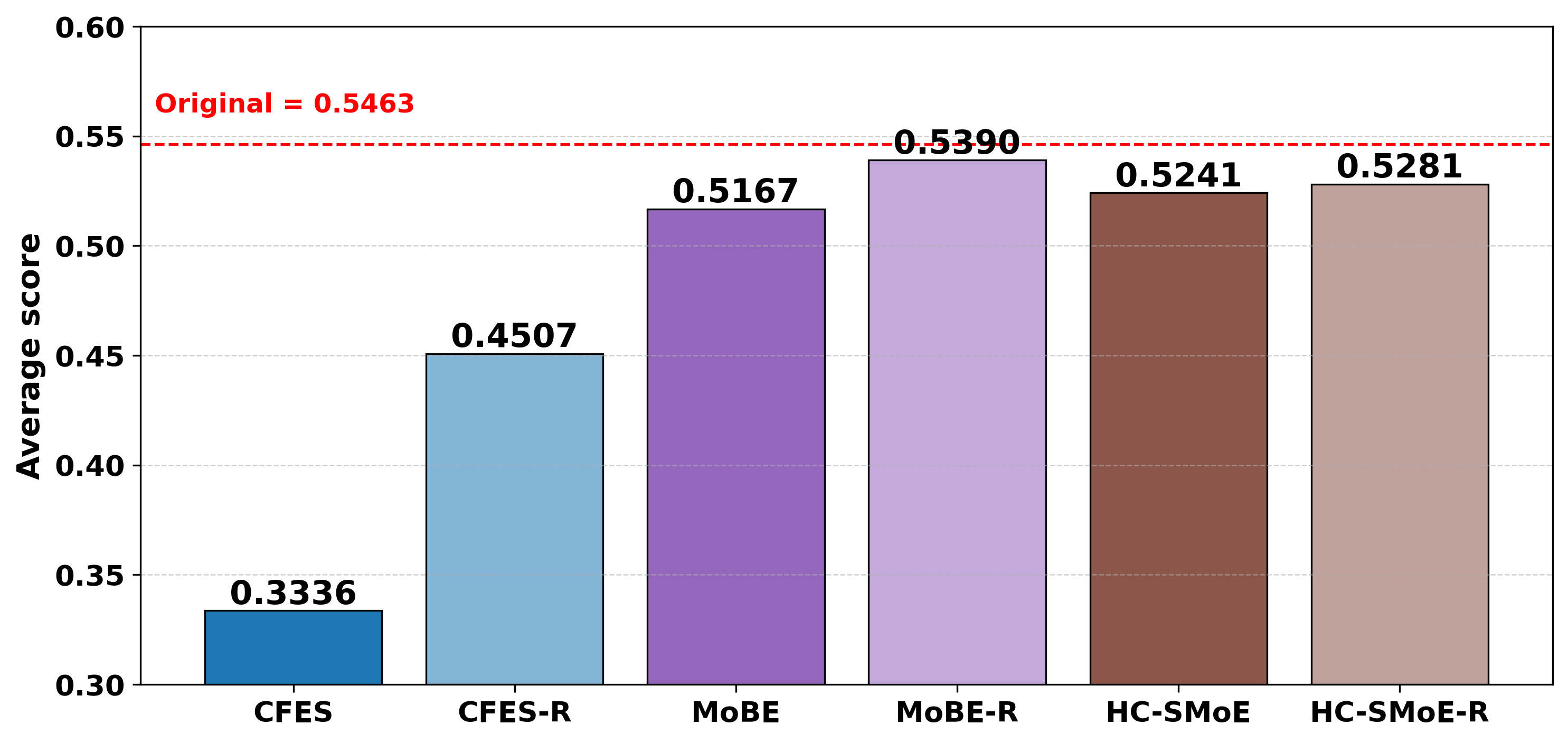}
    \caption{Robustness of Router KD under Milder Compression (75\% Retention). Average benchmark scores for Qwen3 when retaining 75\% of Expert parameters. Comparing standard compression against Router KD (suffix -R) relative to the original model's mean (red line) confirms that Router KD consistently recovers performance across different compression ratios.}
    \label{fig:average} 
\end{figure}


\paragraph{Catastrophic Collapse after Compression}
Additionally, we observed that Router KD yields no performance improvements when the model suffers from catastrophic collapse—that is, when benchmark scores drop to near-zero immediately after compression. This suggests that in catastrophic collapse, the failure likely stems from damage beyond misrouting (e.g., degraded expert representations), which router-only calibration cannot recover.

\paragraph{Performance Gains after Compression}
Furthermore, upon investigating cases where Router KD led to performance regression, we found that these often coincided with instances where compression paradoxically improved model performance. While compression generally degrades performance, there are rare instances where it boosts scores in specific benchmarks or domains. In such scenarios, Router KD becomes counterproductive because the Teacher (the original model) effectively underperforms compared to the Student (the compressed model). However, given that the original model typically outperforms the compressed version, such cases are infrequent.

\subsection{Additional Experiments}

We additionally examined whether the calibration effect of Router KD is preserved under different compression ratios. Specifically, to verify that the observed gains are not limited to a particular compression setting (e.g., 62.5\%), we applied a milder compression configuration to Qwen3, retaining 75\% of the total parameters, and then evaluated the effect of Router KD. The experimental results are summarized in Figure~\ref{fig:average}, with detailed numerical values reported in Table~\ref{tab:additional}. Even under this alternative compression ratio, Router KD exhibited pronounced improvements across all benchmark categories and mitigated the performance drop from the original model compared to the non-KD baseline. These results confirm that Router KD provides robust and consistent recovery effects across compression ratios, rather than being effective only at a specific pruning level.

\section{Conclusion}

In this work, we systematized MoE compression into Expert Pruning, Editing, and Merging, identifying router-expert mismatch as a primary cause of performance degradation. We introduced Router Knowledge Distillation (Router KD), a lightweight strategy that calibrates only the router parameters to mitigate this misalignment. Empirical results confirm that Router KD consistently recovers performance across all paradigms, proving particularly effective for fine-grained architectures with complex routing spaces. We conclude that efficient MoE compression requires accompanying expert modification with minimal router calibration.

\section*{Impact Statement}

This paper presents work whose goal is to advance the field of Machine Learning, specifically addressing the deployment challenges of Large Language Models (LLMs). By enabling effective compression of Mixture-of-Experts (MoE) architectures without the need for resource-intensive full retraining, our proposed method, Router Knowledge Distillation, significantly reduces the memory footprint and computational costs required for inference. This has two primary societal implications: first, it contributes to environmental sustainability by lowering the energy consumption and carbon footprint associated with running large-scale models. Second, it facilitates the democratization of AI by making powerful foundation models accessible to researchers and practitioners with limited hardware resources (e.g., consumer-grade GPUs), thereby reducing the barrier to entry in the field. While we acknowledge the general ethical risks associated with LLMs, such as potential bias or misuse, our specific algorithmic contribution primarily focuses on efficiency and does not introduce new negative societal consequences.

\nocite{langley00}

\bibliography{paper_reference}
\bibliographystyle{icml2026}

\newpage
\appendix
\onecolumn

\section{Original MoE (before Compression)}
\label{appendix:equations1}

In the original MoE model before compression, an input ($ x $) first passes through the gate network, producing $n$ \textit{expert activation scores}. Depending on the implementation, the gate network outputs either the raw logits or the probabilities after a softmax operation. Let these $n$ \textit{expert activation scores} be denoted as ($ g_0, \ldots, g_n $). Likewise, let the $n$ experts in the same layer as the gate network be represented as ($ E_0, \ldots, E_n $).

Assume that this MoE model activates the top-$k$ experts. In this case, the computation for the input $x$ is performed as follows.

$$
\mathcal{S} \subset \{0, 1, \ldots, n-1\}, \quad |S| = k, \quad k < n
$$

We define the renormalized expert activation weights as
$$
\tilde{g}_i = \frac{g_i}{\sum_{j \in \mathcal{S}} g_j}, \quad i \in \mathcal{S}
$$

Using these normalized weights, the output of the MoE layer for input ($x$) is computed as:
$$
y = \sum_{i \in \mathcal{S}} \tilde{g}_i \cdot E_i(x).
$$

\section{Expert Pruning ($N \rightarrow N-\alpha$)}
\label{appendix:equations_pruning}

When the original MoE model undergoes pruning, the inference process can fall into one of the following three scenarios:

1. Best scenario: All experts selected by the original model remain available after pruning, and the same experts are used without any change.

2. Most common scenario: Pruning is imperfect, and while some originally selected experts remain available and are used as before, others are removed. For the dropped experts, the model must instead rely on alternative experts as substitutes.

3. Worst scenario: All experts that were originally selected are pruned out, forcing the model to replace every originally chosen expert with different ones.

Let

\begin{itemize}
    \item $\mathcal{S}$ be the set of expert indices selected by the original MoE model,
    \item $\mathcal{P} \subseteq \{0, \ldots, n-1\}$ be the set of experts that remain after pruning, and $|\mathcal{S}| \leq |\mathcal{P}|$.
    \item $\mathcal{S}'$ be the set of experts effectively used by the pruned model.
\end{itemize}
We analyze the relationship between $\mathcal{S}$, $\mathcal{P}$ and $\mathcal{S}'$.

\noindent\textbf{Gate scores before and after pruning.}
We denote by $g_i^{\mathrm{orig}}(x)$ the expert activation scores produced by the original gate network for an input $x$, and by $g_i^{\mathrm{pruned}}(x)$ the scores produced when running the pruned model end-to-end.
For any index set $\mathcal{A}$, we define the corresponding renormalized weights as
$$
\tilde{g}_i^{\mathrm{orig},(\mathcal{A})}(x)
= \frac{g_i^{\mathrm{orig}}(x)}{\sum_{j \in \mathcal{A}} g_j^{\mathrm{orig}}(x)} 
$$

$$
\tilde{g}_i^{\mathrm{pruned},(\mathcal{A})}(x)
= \frac{g_i^{\mathrm{pruned}}(x)}{\sum_{j \in \mathcal{A}} g_j^{\mathrm{pruned}}(x)},
\quad i \in \mathcal{A}.
$$

\subsection{Best Scenario}

In the best scenario, all originally selected experts remain after pruning:

$
\mathcal{S} \subseteq \mathcal{P}
\quad\Longrightarrow\quad
\mathcal{S}' = \mathcal{S}.
$

The original and pruned MoE outputs for input $x$ can then be written as
$$
y_{\mathrm{orig}}(x)
= \sum_{i \in \mathcal{S}} \tilde{g}_i^{\mathrm{orig},(\mathcal{S})}(x)\, E_i(x)
$$

$$
y_{\mathrm{pruned}}^{\mathrm{best}}(x)
= \sum_{i \in \mathcal{S}} \tilde{g}_i^{\mathrm{pruned},(\mathcal{S})}(x)\, E_i(x).
$$

The difference between the original and pruned MoE outputs in the best scenario is then given by:

$$
\begin{aligned}
& \left\|y_{\mathrm{orig}}(x) - y_{\mathrm{pruned}}^{\mathrm{best}}(x) \right\|=\\
& \left\|\sum_{i \in \mathcal{S}} \tilde{g}_i^{\mathrm{orig},(\mathcal{S})}(x)\, E_i(x) - \sum_{i \in \mathcal{S}} \tilde{g}_i^{\mathrm{pruned},(\mathcal{S})}(x)\, E_i(x)\right\|=\\
& \left\| \sum_{i \in \mathcal{S}} \left( \tilde{g}_i^{\mathrm{orig},(\mathcal{S})}(x) - \tilde{g}_i^{\mathrm{pruned},(\mathcal{S})}(x) \right) E_i(x) \right\|
\end{aligned}
$$

In this scenario, the set of active experts is identical to that of the original model. However, note that even if the selected experts match exactly pre- and post-pruning, the router's output values are unlikely to be identical. Because MoE LLMs are multi-layered, for the router outputs to be perfectly identical, this best-case scenario must be satisfied in every layer, which is statistically improbable. Therefore, it can be trivially stated that $\tilde{g}_i^{\mathrm{orig},(\mathcal{A})}(x) \neq \tilde{g}_i^{\mathrm{pruned},(\mathcal{A})}(x)$

\subsection{Most Common Scenario (Partial Overlap)}

Some selected experts survive, but some are removed:

$
\emptyset \neq \mathcal{S} \cap \mathcal{P} \neq \mathcal{S}.
$

Let the surviving experts be:

$
\mathcal{T} = \mathcal{S} \cap \mathcal{P}.
$

Let the dropped experts be:

$
\mathcal{D} = \mathcal{S} \setminus \mathcal{P}.
$

Then the pruned model must select replacement experts for $\mathcal{D}$,
forming:

$\mathcal{S}' = \mathcal{T} \cup \mathcal{R}$ where $\mathcal{R} \subseteq \mathcal{P} \setminus \mathcal{S}$.

In this case, the original MoE output is
$$
y_{orig}(x) = \sum_{i \in \mathcal{S}} \tilde{g}_i^{orig, (\mathcal{S})}(x) E_i(x)
$$
while the pruned model output decomposes as
$$
y_{pruned}^{common}(x) = \sum_{i \in \mathcal{T}} \tilde{g}_i^{pruned, (\mathcal{S}')}(x) E_i(x) + \sum_{i \in \mathcal{R}} \tilde{g}_i^{pruned, (\mathcal{S}')}(x) E_i(x)
$$

where the first term corresponds to the experts shared with the original model and the second term to the substituted experts.

\subsection{Worst Scenario}

All originally selected experts are pruned out:

$
\mathcal{S} \cap \mathcal{P} = \emptyset.
$

Thus, all experts must be substituted:

$
\mathcal{S}' \subseteq \mathcal{P}, \qquad
|\mathcal{S}'| = k, \qquad
\mathcal{S}' \cap \mathcal{S} = \emptyset.
$

In this case, the original MoE output is
$$
y_{orig}(x) = \sum_{i \in \mathcal{S}} \tilde{g}_i^{orig, (\mathcal{S})}(x) E_i(x)
$$
while the pruned model output becomes
$$
y_{pruned}^{worst}(x) = \sum_{i \in \mathcal{S}'} \tilde{g}_i^{pruned, (\mathcal{S}')}(x) E_i(x)
$$


\subsection{Difference between original and pruned MoE outputs} 

\subsubsection{Best Scenario}
In this scenario, although the set of selected experts remains identical ($\mathcal{S}' = \mathcal{S}$), the output discrepancy is not zero. This error is solely attributable to the deviation in normalized router scores ($\tilde{g}^{\text{orig}}$ vs. $\tilde{g}^{\text{pruned}}$). It highlights that even in the ideal case of expert retention, router calibration is necessary to align the weighting distribution.
$$
\begin{aligned}
& \left\|y_{\mathrm{orig}}(x) - y_{\mathrm{pruned}}^{\mathrm{best}}(x) \right\|=\\
& \left\|\sum_{i \in \mathcal{S}} \tilde{g}_i^{\mathrm{orig},(\mathcal{S})}(x)\, E_i(x) - \sum_{i \in \mathcal{S}} \tilde{g}_i^{\mathrm{pruned},(\mathcal{S})}(x)\, E_i(x)\right\|=\\
& \left\| \sum_{i \in \mathcal{S}} \left( \tilde{g}_i^{\mathrm{orig},(\mathcal{S})}(x) - \tilde{g}_i^{\mathrm{pruned},(\mathcal{S})}(x) \right) E_i(x) \right\|
\end{aligned}
$$

\subsubsection{Most Common Scenario (Partial Overlap)}
Here, the total output difference decomposes into three distinct components:
\begin{itemize}
    \item \textbf{Weight Shift ($\mathcal{T}$)}: The discrepancy caused by altered router weights on the shared experts.
    \item \textbf{Information Loss ($\mathcal{D}$)}: The contribution of the original experts that were dropped, representing missing knowledge.
    \item \textbf{Substitution Noise  ($\mathcal{R}$)}: The impact of newly activated experts that were not selected by the original model, potentially introducing distributional shifts.
\end{itemize}
This formulation clearly shows that the error is driven not just by which experts are lost, but also by how the router re-distributes probability mass among the remaining and new experts.
$$
\begin{aligned}
& \left\|y_{\mathrm{orig}}(x) - y_{\mathrm{pruned}}^{\mathrm{common}}(x) \right\|=\\
& \left\|\sum_{i \in \mathcal{S}} \tilde{g}_i^{\mathrm{orig},(\mathcal{S})}(x)\, E_i(x) - \left( \sum_{i \in \mathcal{T}} \tilde{g}_i^{\mathrm{pruned},(\mathcal{S}')}(x)\, E_i(x)  + \sum_{i \in \mathcal{R}} \tilde{g}_i^{\mathrm{pruned},(\mathcal{S}')}(x)\, E_i(x)\right)\right\| =\\
& \left\| \sum_{i \in \mathcal{T}} \left( \tilde{g}_i^{\mathrm{orig},(\mathcal{S})}(x) - \tilde{g}_i^{\mathrm{pruned},(\mathcal{S}')}(x) \right) E_i(x) + \sum_{i \in \mathcal{D}} \tilde{g}_i^{\mathrm{orig},(\mathcal{S})}(x)\, E_i(x) - \sum_{i \in \mathcal{R}} \tilde{g}_i^{\mathrm{pruned},(\mathcal{S}')}(x)\, E_i(x) \right\|
\end{aligned}
$$

\subsubsection{Worst Scenario}

In the worst-case scenario, the sets of active experts are disjoint ($\mathcal{S} \cap \mathcal{S}' = \emptyset$). Consequently, the model completely loses the original computational path and relies entirely on a different set of experts. This results in the maximum divergence, as the model fails to utilize any of the originally intended parameter knowledge.

$$
\begin{aligned}
& \left\|y_{\mathrm{orig}}(x) - y_{\mathrm{pruned}}^{\mathrm{worst}}(x) \right\|=\\
& \left\|\sum_{i \in \mathcal{S}} \tilde{g}_i^{\mathrm{orig},(\mathcal{S})}(x)\, E_i(x) - \sum_{i \in \mathcal{S}'} \tilde{g}_i^{\mathrm{pruned},(\mathcal{S}')}(x)\, E_i(x)\right\|
\end{aligned}
$$

\section{Expert Editing ($N \rightarrow N$, Parameters $P \rightarrow P'$)}
\label{appendix:equations_editing}

Unlike pruning, Expert Editing preserves the total number of experts. However, as the parameters in preceding layers are modified, the router's computation results may shift. Consequently, for the same input, the edited model may select different experts compared to the original model. The inference process can fall into one of the following three scenarios:

1. Best scenario: The router's selection remains completely unchanged after editing, meaning the exact same experts chosen by the original model are utilized.

2. Most common scenario: Due to the router's altered expert activation scores following the editing process, the selection is partially changed. While some originally selected experts are retained and used as before, others are replaced by different experts.

3. Worst scenario: The router's output shifts drastically such that none of the originally selected experts are chosen, and a completely different set of experts is activated.

Let
\begin{itemize}
    \item $\mathcal{S}$ be the set of expert indices selected by the original MoE model,
    \item $\mathcal{S}^{\mathrm{edit}} \subseteq \{0, \ldots, n-1\}$ be the set of expert indices selected by the edited MoE model, and $|\mathcal{S}| = |\mathcal{S}^{\mathrm{edit}}|$.
    \item $E_i$ denote the original experts, and $X_i$ denote the corresponding edited experts obtained by modifying $E_i$.
\end{itemize}

\noindent\textbf{Gate scores before and after editing.}
As before, we denote by $g_i^{\mathrm{orig}}(x)$ the expert activation scores produced by the original gate network for an input $x$.
We denote by $g_i^{\mathrm{edit}}(x)$ the activation scores produced when running the edited MoE model end-to-end on the same input.
For any index set $\mathcal{A}$, we define the renormalized weights for the edited model as
$$
\tilde{g}_i^{\mathrm{edit},(\mathcal{A})}(x)
= \frac{g_i^{\mathrm{edit}}(x)}{\sum_{j \in \mathcal{A}} g_j^{\mathrm{edit}}(x)},
\quad i \in \mathcal{A}.
$$

\subsection{Best Scenario}

In the best scenario, the router's top-$k$ selection remains unchanged even after expert editing, i.e.,
$$
\mathcal{S}^{\mathrm{edit}} = \mathcal{S}.
$$
The original MoE output for input $x$ is
$$
y_{\mathrm{orig}}(x)
= \sum_{i \in \mathcal{S}} \tilde{g}_i^{\mathrm{orig},(\mathcal{S})}(x)\, E_i(x),
$$
while the edited MoE output in the best scenario becomes
$$
y_{\mathrm{edit}}^{\mathrm{best}}(x)
= \sum_{i \in \mathcal{S}} \tilde{g}_i^{\mathrm{edit},(\mathcal{S})}(x)\, X_i(x).
$$
Even though the index set of selected experts is identical, both the gate scores and the expert functions may differ between $y_{\mathrm{orig}}(x)$ and $y_{\mathrm{edit}}^{\mathrm{best}}(x)$.

The difference between the original and edited MoE outputs in the best scenario is then given by:

$$
\begin{aligned}
& \left\|y_{\mathrm{orig}}(x) - y_{\mathrm{edit}}^{\mathrm{best}}(x) \right\|=\\
& \left\|\sum_{i \in \mathcal{S}} \tilde{g}_i^{\mathrm{orig},(\mathcal{S})}(x)\, E_i(x) - \sum_{i \in \mathcal{S}} \tilde{g}_i^{\mathrm{edit},(\mathcal{S})}(x)\, X_i(x)\right\| \\
& \left\|\sum_{i \in \mathcal{S}} \left( \tilde{g}_i^{\mathrm{orig},(\mathcal{S})}(x)\, E_i(x) - \tilde{g}_i^{\mathrm{edit},(\mathcal{S})}(x)\, X_i(x)\right)\right\|
\end{aligned}
$$

Even if sophisticated mathematical approximation techniques render $E_i$ and $X_i$ nearly identical, they are not strictly identical; thus, $g_i^{\mathrm{orig}}$ and $g_i^{\mathrm{edit}}$ cannot be equal. Given that each expert is a matrix containing over 1 million parameters, even slight deviations in router outputs—despite the similarity between $E_i$ and $X_i$—inevitably lead to non-negligible differences in the final output.

\subsection{Most Common Scenario (Partial Overlap)}
In this scenario, the router's selection is partially preserved but also altered due to the editing process. We define the sets of shared ($\mathcal{T}$), dropped ($\mathcal{D}$), and newly introduced ($\mathcal{R}$) experts as:
$$
\mathcal{T} = \mathcal{S} \cap \mathcal{S}^{edit}, \quad \mathcal{D} = \mathcal{S} \setminus \mathcal{S}^{edit}, \quad \mathcal{R} = \mathcal{S}^{edit} \setminus \mathcal{S}
$$
The effective set of experts for the edited model is formed by $\mathcal{S}^{edit} = \mathcal{T} \cup \mathcal{R}$.
In this case, the original MoE output is
$$
y_{\mathrm{orig}}(x) = \sum_{i \in \mathcal{S}} \tilde{g}_i^{\mathrm{orig}, (\mathcal{S})}(x) E_i(x)
$$
while the edited model output decomposes as
$$
y_{\mathrm{edit}}^{\mathrm{common}}(x) = \sum_{i \in \mathcal{T}} \tilde{g}_i^{\mathrm{edit}, (\mathcal{S}^{edit})}(x) X_i(x) + \sum_{i \in \mathcal{R}} \tilde{g}_i^{\mathrm{edit}, (\mathcal{S}^{edit})}(x) X_i(x)
$$
where the first term represents the edited versions of the originally selected experts, and the second term represents the newly activated experts.


\subsection{Worst Scenario}
In the worst-case scenario, the router's behavior shifts drastically such that there is no overlap between the original and edited selections ($\mathcal{S} \cap \mathcal{S}^{edit} = \emptyset$).
Consequently, the model relies entirely on a disjoint set of experts.
The original output is
$$
y_{\mathrm{orig}}(x) = \sum_{i \in \mathcal{S}} \tilde{g}_i^{\mathrm{orig}, (\mathcal{S})}(x) E_i(x)
$$
whereas the edited model output becomes
$$
y_{\mathrm{edit}}^{\mathrm{worst}}(x) = \sum_{i \in \mathcal{S}^{edit}} \tilde{g}_i^{\mathrm{edit}, (\mathcal{S}^{edit})}(x) X_i(x)
$$
Here, both the selected expert indices and the underlying expert parameters differ entirely from the original model.
\newpage

\subsection{Difference between original and edited MoE outputs} 

\subsubsection{Best Scenario}
Even in the best-case scenario where the router selects the exact same expert indices ($\mathcal{S}^{edit} = \mathcal{S}$), the discrepancy arises from two sources: (1) the approximation error of the experts themselves ($E_i \to X_i$) and (2) the shift in router weights ($\tilde{g}^{orig} \to \tilde{g}^{edit}$).

$$
\begin{aligned}
& \left\|y_{\mathrm{orig}}(x) - y_{\mathrm{edit}}^{\mathrm{best}}(x) \right\| \\
&= \left\|\sum_{i \in \mathcal{S}} \tilde{g}_i^{\mathrm{orig},(\mathcal{S})}(x)\, E_i(x) - \sum_{i \in \mathcal{S}} \tilde{g}_i^{\mathrm{edit},(\mathcal{S})}(x)\, X_i(x)\right\| \\
&= \left\|\sum_{i \in \mathcal{S}} \left( \tilde{g}_i^{\mathrm{orig},(\mathcal{S})}(x)\, E_i(x) - \tilde{g}_i^{\mathrm{edit},(\mathcal{S})}(x)\, X_i(x)\right)\right\|
\end{aligned}
$$

\subsubsection{Most Common Scenario (Partial Overlap)}
The total output difference decomposes into three components, reflecting both structural changes in the router and parameter approximation in the experts:
\begin{itemize}
    \item \textbf{Approximation \& Weight Shift ($\mathcal{T}$):} The compound error arising from the parameter compression of shared experts ($E_i \to X_i$) and the deviation in their router weights.
    \item \textbf{Information Loss ($\mathcal{D}$):} The loss of knowledge from the original experts ($E_i$) that were dropped.
    \item \textbf{Substitution Noise ($\mathcal{R}$):} The impact of newly introduced, compressed experts ($X_i$) that were not originally selected.
\end{itemize}
$$
\begin{aligned}
& \left\|y_{\mathrm{orig}}(x) - y_{\mathrm{edit}}^{\mathrm{common}}(x) \right\|\\
&= \left\|\sum_{i \in \mathcal{S}} \tilde{g}_i^{\mathrm{orig},(\mathcal{S})}(x)\, E_i(x) - \left(\sum_{i \in \mathcal{T}} \tilde{g}_i^{\mathrm{edit},(\mathcal{S}^{\mathrm{edit}})}(x)\, X_i(x)  + \sum_{i \in \mathcal{R}} \tilde{g}_i^{\mathrm{edit},(\mathcal{S}^{\mathrm{edit}})}(x)\, X_i(x)\right)\right\| \\
&= \left\| \sum_{i \in \mathcal{T}} \left( \tilde{g}_i^{\mathrm{orig},(\mathcal{S})}(x)\, E_i(x) - \tilde{g}_i^{\mathrm{edit},(\mathcal{S}^{\mathrm{edit}})}(x)\, X_i(x) \right) \right. \left. + \sum_{i \in \mathcal{D}} \tilde{g}_i^{\mathrm{orig},(\mathcal{S})}(x)\, E_i(x) - \sum_{i \in \mathcal{R}} \tilde{g}_i^{\mathrm{edit},(\mathcal{S}^{\mathrm{edit}})}(x)\, X_i(x) \right\|
\end{aligned}
$$

\subsubsection{Worst Scenario}
With disjoint active sets ($\mathcal{S} \cap \mathcal{S}^{edit} = \emptyset$), the model relies entirely on a different set of compressed experts ($X_i$). This maximizes divergence as the original parameters are completely abandoned.
$$
\begin{aligned}
\left\|y_{\mathrm{orig}}(x) - y_{\mathrm{edit}}^{\mathrm{worst}}(x) \right\| &= \left\|\sum_{i \in \mathcal{S}} \tilde{g}_i^{\mathrm{orig},(\mathcal{S})}(x) E_i(x) - \sum_{i \in \mathcal{S}^{edit}} \tilde{g}_i^{\mathrm{edit},(\mathcal{S}^{edit})}(x) X_i(x)\right\|
\end{aligned}
$$

\newpage

\section{Expert Merging ($N \rightarrow M$, where $M < N$)}
\label{appendix:equations_merging}

Expert Merging is primarily implemented by replacing the original experts with a merged representation to reduce the memory footprint. Structurally, multiple original experts are clustered into a single representative expert. For instance, if Expert 0 and Expert 1 are merged, they share the same parameters in the compressed model, effectively mapping multiple original indices to a single merged index.

To formalize this, let $N$ be the number of original experts and $M$ be the number of merged experts ($M < N$). We define a surjective mapping function $\phi: \{1, \dots, N\} \to \{1, \dots, M\}$ that assigns each original expert index $i$ to a merged expert cluster index $c$. Consequently, the output of the merged model is computed using the merged parameters $M_c$:
$$
M_c \approx E_i, \quad \forall i \text{ such that } \phi(i) = c
$$

\subsection{Set Definitions and Output Formulation}

Since the original and merged models operate in different index spaces, direct comparison of selected indices is not possible. We analyze the discrepancy by projecting the original selection into the merged cluster space.

Let $\mathcal{S}$ be the set of indices selected by the original router, where $|\mathcal{S}| = k$. We define the \textbf{Projected Original Set} $\mathcal{C}_{\text{proj}}$ as the set of unique clusters required by the original selection:
$$
\mathcal{C}_{\text{proj}} = \{ \phi(i) \mid i \in \mathcal{S} \}
$$
Let $\mathcal{S}^{merge}$ be the set of indices selected by the merged model's router, where $|\mathcal{S}^{merge}| = k_{merge}$. We define the intersection ($\mathcal{T}$), dropped ($\mathcal{D}$), and newly introduced ($\mathcal{R}$) sets within the cluster space:

\begin{itemize}
    \item \textbf{Shared Clusters ($\mathcal{T} = \mathcal{C}_{\text{proj}} \cap \mathcal{S}^{merge}$):} Clusters intended by the original model and selected by the merged model.
    \item \textbf{Dropped Clusters ($\mathcal{D} = \mathcal{C}_{\text{proj}} \setminus \mathcal{S}^{merge}$):} Clusters required by the original input but missed by the merged router.
    \item \textbf{Substituted Clusters ($\mathcal{R} = \mathcal{S}^{merge} \setminus \mathcal{C}_{\text{proj}}$):} Clusters selected by the merged router that were not implied by the original selection.
\end{itemize}

Based on these definitions, we analyze the scenarios below based on the relationship between the number of required clusters ($|\mathcal{C}_{\text{proj}}|$) and the merged router's capacity ($k_{merge}$).

\newpage

\subsection{Case 1: All Original Experts Co-located in One Cluster}
This case occurs when all selected original experts map to a single merged cluster ($|\mathcal{C}_{\text{proj}}| = 1$). Let $c^*$ be that unique cluster index.

\paragraph{Scenario 1.1: Best Case.}
The merged router selects exactly the cluster $c^*$.
$$ \mathcal{S}^{merge} = \{c^*\} $$
$$ y_{\mathrm{merge}}^{(1.1)}(x) = \tilde{g}_{c^*}^{\mathrm{merge}}(x) M_{c^*}(x) $$

\paragraph{Scenario 1.2: Most Common (Partial Noise).}
The router selects the correct cluster $c^*$ but also activates irrelevant clusters ($\mathcal{R}$).
$$ c^* \in \mathcal{S}^{merge}, \quad \mathcal{R} = \mathcal{S}^{merge} \setminus \{c^*\} \neq \emptyset $$
$$ y_{\mathrm{merge}}^{(1.2)}(x) = \tilde{g}_{c^*}^{\mathrm{merge}}(x) M_{c^*}(x) + \sum_{c \in \mathcal{R}} \tilde{g}_c^{\mathrm{merge}}(x) M_c(x) $$

\paragraph{Scenario 1.3: Worst Case (Miss).}
The router completely misses the required cluster $c^*$.
$$ c^* \notin \mathcal{S}^{merge} $$
$$ y_{\mathrm{merge}}^{(1.3)}(x) = \sum_{c \in \mathcal{S}^{merge}} \tilde{g}_c^{\mathrm{merge}}(x) M_c(x) $$

\subsection{Case 2: Distributed Experts (Within Capacity)}
Here, the original experts map to multiple distinct clusters, but the number of required clusters is within the merged router's selection capacity ($1 < |\mathcal{C}_{\text{proj}}| \le k_{merge}$).

\paragraph{Scenario 2.1: Best Case (Perfect Match).}
The router selects exactly the set of projected clusters.
$$ \mathcal{S}^{merge} = \mathcal{C}_{\text{proj}} $$
$$ y_{\mathrm{merge}}^{(2.1)}(x) = \sum_{c \in \mathcal{C}_{\text{proj}}} \tilde{g}_c^{\mathrm{merge}}(x) M_c(x) $$

\paragraph{Scenario 2.2: Most Common (Partial Overlap).}
Some correct clusters are selected ($\mathcal{T}$), some are missed ($\mathcal{D}$), and noise is added ($\mathcal{R}$).
$$ \mathcal{T} = \mathcal{C}_{\text{proj}} \cap \mathcal{S}^{merge} \neq \emptyset, \quad \mathcal{D} = \mathcal{C}_{\text{proj}} \setminus \mathcal{T} \neq \emptyset, \quad \mathcal{R} = \mathcal{S}^{merge} \setminus \mathcal{T} \neq \emptyset $$
$$ y_{\mathrm{merge}}^{(2.2)}(x) = \sum_{c \in \mathcal{T}} \tilde{g}_c^{\mathrm{merge}}(x) M_c(x) + \sum_{c \in \mathcal{R}} \tilde{g}_c^{\mathrm{merge}}(x) M_c(x) $$

\paragraph{Scenario 2.3: Worst Case (Disjoint).}
No correct clusters are selected.
$$ \mathcal{C}_{\text{proj}} \cap \mathcal{S}^{merge} = \emptyset $$
$$ y_{\mathrm{merge}}^{(2.3)}(x) = \sum_{c \in \mathcal{S}^{merge}} \tilde{g}_c^{\mathrm{merge}}(x) M_c(x) $$

\subsection{Case 3: Over-Distributed (Capacity Exceeded)}
This critical scenario occurs when the original experts are scattered across more clusters than the merged router is allowed to select ($|\mathcal{C}_{\text{proj}}| > k_{merge}$). This implies inevitable structural information loss.

\paragraph{Scenario 3.1: Best Possible (Capacity Saturation).}
Even in the best case, the router can only select a subset of the required clusters. Let $\mathcal{T}_{max} \subset \mathcal{C}_{\text{proj}}$ be the largest possible subset ($|\mathcal{T}_{max}| = k_{merge}$).
$$ \mathcal{S}^{merge} = \mathcal{T}_{max}, \quad \mathcal{D}_{inevitable} = \mathcal{C}_{\text{proj}} \setminus \mathcal{T}_{max} \neq \emptyset $$
$$ y_{\mathrm{merge}}^{(3.1)}(x) = \sum_{c \in \mathcal{T}_{max}} \tilde{g}_c^{\mathrm{merge}}(x) M_c(x) $$

\paragraph{Scenario 3.2: Most Common (Sub-optimal Selection).}
The router selects fewer correct clusters than its capacity allows, or picks wrong ones.
$$ \mathcal{T} = \mathcal{C}_{\text{proj}} \cap \mathcal{S}^{merge}, \quad |\mathcal{T}| < k_{merge}, \quad \mathcal{R} = \mathcal{S}^{merge} \setminus \mathcal{T} \neq \emptyset $$
$$ y_{\mathrm{merge}}^{(3.2)}(x) = \sum_{c \in \mathcal{T}} \tilde{g}_c^{\mathrm{merge}}(x) M_c(x) + \sum_{c \in \mathcal{R}} \tilde{g}_c^{\mathrm{merge}}(x) M_c(x) $$

\paragraph{Scenario 3.3: Worst Case.}
The router selects clusters completely disjoint from $\mathcal{C}_{\text{proj}}$.
$$ \mathcal{C}_{\text{proj}} \cap \mathcal{S}^{merge} = \emptyset $$
$$ y_{\mathrm{merge}}^{(3.3)}(x) = \sum_{c \in \mathcal{S}^{merge}} \tilde{g}_c^{\mathrm{merge}}(x) M_c(x) $$

\subsection{Difference between original and merged MoE outputs}

We analyze the output discrepancy $\|y_{\mathrm{orig}}(x) - y_{\mathrm{merge}}(x)\|$ across the scenarios defined in previous sections.

\subsubsection*{Case 1: All originally selected experts belong to the same cluster}

\textbf{Scenario 1.1: Best Case} \\
The router correctly identifies the unique cluster $c^\star$. The error arises solely from the \textbf{Merging Approximation}.
$$
\begin{aligned}
& \left\|y_{\mathrm{orig}}(x) - y_{\mathrm{merge}}^{(1.1)}(x) \right\| \\
&= \left\|
    \sum_{i \in \mathcal{S}} 
      \tilde{g}_i^{\mathrm{orig},(\mathcal{S})}(x)\, E_i(x)
    - \tilde{g}_{c^*}^{\mathrm{merge}}(x)\, M_{c^*}(x)
   \right\|.
\end{aligned}
$$

\textbf{Scenario 1.2: Most Common (Partial Noise)} \\
The router activates irrelevant clusters ($\mathcal{R}$). The error decomposes into \textbf{Merging Approximation} and \textbf{Substitution Noise}.
$$
\begin{aligned}
& \left\|y_{\mathrm{orig}}(x) - y_{\mathrm{merge}}^{(1.2)}(x) \right\| \\
&= \left\|
    \sum_{i \in \mathcal{S}} 
      \tilde{g}_i^{\mathrm{orig},(\mathcal{S})}(x)\, E_i(x)
    - \left(
        \tilde{g}_{c^*}^{\mathrm{merge}}(x)\, M_{c^*}(x)
        + \sum_{c \in \mathcal{R}}
          \tilde{g}_c^{\mathrm{merge}}(x)\, M_c(x)
      \right)
   \right\|.
\end{aligned}
$$

\textbf{Scenario 1.3: Worst Case (Miss)} \\
Complete \textbf{Information Loss} of the original knowledge, replaced entirely by \textbf{Substitution Noise}.
$$
\begin{aligned}
& \left\|y_{\mathrm{orig}}(x) - y_{\mathrm{merge}}^{(1.3)}(x) \right\| \\
&= \left\|
    \sum_{i \in \mathcal{S}} 
      \tilde{g}_i^{\mathrm{orig},(\mathcal{S})}(x)\, E_i(x)
    - \sum_{c \in \mathcal{S}^{merge}}
          \tilde{g}_c^{\mathrm{merge}}(x)\, M_c(x)
   \right\|.
\end{aligned}
$$

\subsubsection*{Case 2: Originally selected experts share some clusters (Mixed)}

\textbf{Scenario 2.1: Best Case (Perfect Match)} \\
The discrepancy is purely due to \textbf{Merging Approximation} across the active clusters $\mathcal{C}_{\text{proj}}$.
$$
\begin{aligned}
& \left\|y_{\mathrm{orig}}(x) - y_{\mathrm{merge}}^{(2.1)}(x) \right\| \\
&= \left\|
    \sum_{i \in \mathcal{S}} 
      \tilde{g}_i^{\mathrm{orig},(\mathcal{S})}(x)\, E_i(x)
    - \sum_{c \in \mathcal{C}_{\text{proj}}}
         \tilde{g}_c^{\mathrm{merge}}(x)\, M_c(x)
   \right\|.
\end{aligned}
$$

\textbf{Scenario 2.2: Most Common (Partial Overlap)} \\
This involves: \textbf{Merging Approximation ($\mathcal{T}$)}, \textbf{Information Loss ($\mathcal{D}$)}, and \textbf{Substitution Noise ($\mathcal{R}$)}.
$$
\begin{aligned}
& \left\|y_{\mathrm{orig}}(x) - y_{\mathrm{merge}}^{(2.2)}(x) \right\| \\
&= \left\|
    \sum_{i \in \mathcal{S}} 
      \tilde{g}_i^{\mathrm{orig},(\mathcal{S})}(x)\, E_i(x)
    - \left(
        \sum_{c \in \mathcal{T}}
          \tilde{g}_c^{\mathrm{merge}}(x)\, M_c(x)
        + \sum_{c \in \mathcal{R}}
          \tilde{g}_c^{\mathrm{merge}}(x)\, M_c(x)
      \right)
   \right\| \\
&= \left\|
    \left(
        \sum_{i \in \mathcal{S}, \phi(i) \in \mathcal{T}} \tilde{g}_i^{\mathrm{orig}}(x) E_i(x) 
        - \sum_{c \in \mathcal{T}} \tilde{g}_c^{\mathrm{merge}}(x) M_c(x)
    \right)
    + \sum_{i \in \mathcal{S}, \phi(i) \in \mathcal{D}} 
        \tilde{g}_i^{\mathrm{orig}}(x)\, E_i(x)
    - \sum_{c \in \mathcal{R}} 
        \tilde{g}_c^{\mathrm{merge}}(x)\, M_c(x)
   \right\|.
\end{aligned}
$$

\textbf{Scenario 2.3: Worst Case (Disjoint)} \\
Maximum divergence due to complete mismatch ($\mathcal{T} = \emptyset$).
$$
\begin{aligned}
& \left\|y_{\mathrm{orig}}(x) - y_{\mathrm{merge}}^{(2.3)}(x) \right\| \\
&= \left\|
    \sum_{i \in \mathcal{S}} 
      \tilde{g}_i^{\mathrm{orig},(\mathcal{S})}(x)\, E_i(x)
    - \sum_{c \in \mathcal{S}^{\mathrm{merge}}}
        \tilde{g}_c^{\mathrm{merge}}(x)\, M_c(x)
   \right\|.
\end{aligned}
$$

\subsubsection*{Case 3: All originally selected experts belong to distinct clusters}

\textbf{Scenario 3.1: Best Possible (Capacity Saturation)} \\
Even with optimal selection ($\mathcal{T}_{max}$), the model suffers from structural loss ($\mathcal{D}_{inevitable}$).
$$
\begin{aligned}
& \left\|y_{\mathrm{orig}}(x) - y_{\mathrm{merge}}^{(3.1)}(x) \right\| \\
&= \left\|
    \sum_{i \in \mathcal{S}} 
      \tilde{g}_i^{\mathrm{orig},(\mathcal{S})}(x)\, E_i(x)
    - \sum_{c \in \mathcal{T}_{max}}
        \tilde{g}_c^{\mathrm{merge}}(x)\, M_c(x)
   \right\|.
\end{aligned}
$$

\textbf{Scenario 3.2: Most Common (Sub-optimal Selection)} \\
Structurally prone to \textbf{Information Loss ($\mathcal{D}$)} due to limited capacity $k_{merge}$ and sub-optimal selection.
$$
\begin{aligned}
& \left\|y_{\mathrm{orig}}(x) - y_{\mathrm{merge}}^{(3.2)}(x) \right\| \\
&= \left\|
    \sum_{i \in \mathcal{S}} 
      \tilde{g}_i^{\mathrm{orig},(\mathcal{S})}(x)\, E_i(x)
    - \left(
        \sum_{c \in \mathcal{T}}
          \tilde{g}_c^{\mathrm{merge}}(x)\, M_c(x)
        + \sum_{c \in \mathcal{R}}
          \tilde{g}_c^{\mathrm{merge}}(x)\, M_c(x)
      \right)
   \right\|.
\end{aligned}
$$

\textbf{Scenario 3.3: Worst Case} \\
Complete mismatch.
$$
\begin{aligned}
& \left\|y_{\mathrm{orig}}(x) - y_{\mathrm{merge}}^{(3.3)}(x) \right\| \\
&= \left\|
    \sum_{i \in \mathcal{S}} 
      \tilde{g}_i^{\mathrm{orig},(\mathcal{S})}(x)\, E_i(x)
    - \sum_{c \in \mathcal{S}^{\mathrm{merge}}}
        \tilde{g}_c^{\mathrm{merge}}(x)\, M_c(x)
   \right\|.
\end{aligned}
$$

\newpage

\section{Related Works}
\label{appendix:appendix}

Due to the massive VRAM consumption of Large Language Models (LLMs) based on the Mixture-of-Experts (MoE) architecture, deploying these models remains challenging, particularly in resource-constrained environments. Consequently, there is a growing need for MoE compression techniques that minimize the degradation of existing high-performance models while being retraining-free and computationally efficient.
To address this, extensive research has been conducted. In this paper, we propose a taxonomy for MoE compression, categorizing existing methodologies into three classes: \textbf{Expert Pruning}, \textbf{Expert Editing}, and \textbf{Expert Merging}. In this work, we focus exclusively on \textbf{parameter-level} compression. We do not consider bit-level compression of individual parameters (i.e., quantization). In other words, our scope is limited to methods that reduce the number of parameters efficiently, rather than techniques that optimize the bit representation of each parameter, which lies outside the boundaries of this study.

\subsection{Expert Pruning ($N \rightarrow N-\alpha$)}
Expert Pruning reduces the total number of experts by permanently removing $\alpha$ out of the total $N$ experts. Since this method simply drops $\alpha$ experts, the remaining $N-\alpha$ experts are preserved without modification. Consequently, after the parameters are removed, routing is performed exclusively among the remaining $N-\alpha$ experts. The core challenge of this approach lies in identifying the importance of each expert; it is based on the hypothesis that certain experts are less important, redundant, or make negligible contributions to the model's performance, and thus can be removed without significant degradation.

A wide array of methodologies has been proposed in the field of Expert Pruning. NAEE~\cite{NAEE} proposes a method that retains only the combination of experts that minimizes layer-wise reconstruction loss after inferring the MoE LLM with specific calibration data. DiEP~\cite{bai2025diep} approaches the problem by transforming the discrete expert selection task into a differentiable continuous optimization. REAP~\cite{reap} demonstrates that expert merging can lead to functional subspace collapse, resulting in lower performance on text generation tasks compared to pruning; instead, it introduces a pruning criterion that considers both router gate-values and the magnitude of expert outputs. Another study \cite{anonymous2025compressing} mathematically proved that the output discrepancy of the entire model is bounded by the cumulative sum of layer-wise output discrepancies, thereby proposing a layer-wise search instead of a global search. Furthermore, EASY-EP~\cite{easy_ep}, observing that experts in large-scale MoE models are highly specialized and identifiable with limited data, proposes an effective domain-specific pruning method based on expert output magnitude and token variation. MoE Pathfinder~\cite{moepathfinder} evaluates global importance based on the `expert activation trajectory' across all layers to perform pruning. Beyond these specific algorithms, a comprehensive set of evaluation criteria, MC-Suite, was suggested to determine the optimal experts for removal~\cite{mc_suite}, and another study has explored more aggressive approaches, such as removing entire MoE layers or even transformer blocks rather than individual experts~\cite{RS_TMLR}.

\subsection{Expert Editing ($N \rightarrow N$, Parameters $P \rightarrow P'$)}

Expert Editing maintains the total number of experts $N$ but reduces the total number of parameters from $P$ to $P'$ by compressing the internal structure of each expert and adjusting the computation order. This approach is grounded in the hypothesis that the weight matrices ($W$) within experts are over-parameterized and can be approximated by matrices with fewer parameters. The core mechanism involves mathematically decomposing or re-parameterizing the expert matrices. Techniques such as Singular Value Decomposition (SVD), Rank Decomposition, and Tucker Decomposition are employed, and the operation order of the decomposed matrices may be rearranged to maximize efficiency or performance. Furthermore, some approaches attempt to replace expert matrices with smaller, lighter structures, such as vectors.

Expert Editing has recently gained significant attention, with various methodologies being actively proposed. MoE-SVD~\cite{moe-svd} applies SVD to each expert's weight matrix, demonstrating that it is possible to effectively compress experts and reduce parameters while preserving performance. MoLAE~\cite{molae} proposes factorizing an expert into two matrices ($A$ and $B$) using SVD; it designates $B$ as a latent mapping matrix shared across all experts (or expert groups), and $A$ as an expert-specific transformation matrix that operates within the low-dimensional space.
Building on this, MoBE~\cite{mobe} utilizes rank decomposition to decompose experts into a unique matrix $A$ and a shared linear combination of basis matrices. It optimizes these components to minimize the reconstruction error with the original weight matrix, thereby achieving effective compression and performance preservation. Additionally, TD-MoE~\cite{td_moe} treats MoE experts not as independent matrices but as correlated tensors. It aligns the data distribution via whitening and then applies Tucker Decomposition, aiming to capture the correlations among experts through Joint Tensor Decomposition.

\subsection{Expert Merging ($N \rightarrow M$, where $M < N$)}

Expert merging reduces the total number of experts from $N$ to a smaller number $M$ by combining functionally similar experts into synthesized experts. It is motivated by model merging hypotheses~\cite{modelsoup, mergingmodelsfisher}, which posit that when experts exhibit redundancy, their knowledge can be fused to preserve most of the collective capability of the original set. This process often involves (i) identifying redundancy via clustering or similarity matching (e.g., hierarchical clustering or $k$-means) and (ii) fusing experts via parameter- or output-level mechanisms (e.g., weighted averaging or learned linear combinations).

Various methodologies have been proposed for MoE expert merging. HC-SMoE~\cite{hcsmoe} reduces parameters by performing hierarchical clustering based on each expert’s output similarity on calibration data, thereby merging similar experts in a retraining-free manner. PuzzleMoE~\cite{puzzlemoe} is a training-free approach that targets expert pairs by constructing a dual-mask consisting of an entry-wise similarity mask and an activation-weighted saliency mask, enabling selective merging of redundant parameters while preserving expert-specific knowledge. MergeMoE~\cite{mergemoe} reinterprets MoE expert merging from the perspective of output merging rather than parameter averaging, and computes a compression (dimensionality reduction) matrix via least squares based on sample inputs. Complementary to MoE expert merging, the Expert Merging method~\cite{expertmerging} studies merging multiple domain experts (SFT models) by learning layer-wise (and importance-guided chunk-wise) coefficients from unlabeled calibration data to align hidden states and logits across experts.

\subsection{Hybrid Approaches}

Methodologies that combine the three aforementioned approaches are also being actively proposed. These strategies include merging after pruning, editing after pruning, and editing after merging. 

DM-MoE~\cite{drop_or_merge} proposed a `drop-then-merge' hybrid MoE compression method that first prunes (drops) redundant experts and subsequently merges the remaining experts using a graph-based approach. DERN~\cite{dern} compresses MoE models without retraining by pruning redundant experts based on router statistics, decomposing and reallocating the pruned experts into neuron-level segments, and finally merging these segments via clustering. EEP~\cite{eep} introduced a method that reduces inference costs while maintaining or improving performance without retraining; it utilizes a gradient-free evolutionary strategy to prune MoE experts and preserves knowledge through expert merging.

MoNE~\cite{mone} proposes a pruning-and-editing approach that prunes redundant experts and replaces them by editing them into lightweight, input-agnostic `Novice' vectors. MoE-I2~\cite{moei2} suggests a two-stage (Pruning+Decomposition) framework to compress MoE models. It performs non-uniform Inter-Expert Pruning based on layer/expert importance analysis, further compresses the remaining experts via non-uniform Low-Rank Decomposition, and recovers performance using LoRA fine-tuning. 
$D^{2}$-MoE~\cite{d2moe} proposes a delta-based compression method that constructs a shared base weight via Fisher-weighted merging and stores the difference of each expert as a low-rank edit using truncation-aware SVD. MC-SMoE~\cite{mcsmoe} employs M-SMoE to capture expert redundancy using routing policy statistics, align neurons via permutation alignment, and integrate experts through activation frequency-based weighted averaging. Subsequently, it further compresses the integrated experts via low-rank decomposition to maximize memory and parameter efficiency.

Since each individual method possesses distinct advantages and limitations, hybridizing two or more of these approaches represents a promising direction for future research.

\subsection{Importance Of Router}

MoE compression is often framed as modifying expert parameters (e.g., pruning, merging, editing, or quantization), but recent studies consistently highlight that the \emph{router} is a disproportionately high-leverage component: even small perturbations to expert weights or numerics can shift expert outputs, which in turn changes token-to-expert assignments and cascades into larger performance drops. This phenomenon is particularly evident in low-bit quantization. EAC-MoE attributes a major failure mode of quantized MoEs to \emph{expert-shift}—a distortion of expert selection after quantization—and proposes router calibration to mitigate the cumulative accumulation of expert selection shift across layers, using a TopK-MSE objective that focuses alignment on the experts most likely to be selected \cite{EAC_MoE}. Similarly, GEMQ observes that quantization substantially distorts router behavior and shows that merely reusing full-precision router signals is insufficient; instead, a lightweight \emph{global router fine-tuning} step (updating only router parameters) provides substantial gains in perplexity and downstream accuracy under low-bit regimes \cite{anonymous2026towards} . Beyond quantization, Router-Tuning further demonstrates that training only a lightweight routing module can effectively steer computation by deciding when to skip modules, and the approach can be deployed on MoE backbones as well, reinforcing that routing adaptation alone can be impactful \cite{he-etal-2025-router}.

While these works mainly focus on \emph{bit-level} compression (mixed-precision quantization) or \emph{compute-efficient} routing, our setting is complementary: we study \emph{parameter-level} MoE compression (e.g., expert pruning, merging, and editing), where experts are structurally modified or removed. In this regime, the pre-trained router is often no longer well-matched to the post-compression expert set, making router–expert mismatch a key driver of performance degradation. Our findings therefore support a broader conclusion: effective MoE compression requires not only modifying experts, but also \emph{calibrating the router} to remain consistent with the altered expert landscape.

\section{Experiment Settings}
\label{appendix:expsetting}

\subsection{Common protocol in Qwen3.}
For all experiments on the Qwen3 backbone, we followed the default configurations provided by each baseline’s \emph{official} code release as closely as possible.
We only introduced minimal, unavoidable changes to accommodate our server constraints (e.g., the number of GPUs and distributed/runtime configuration), while keeping all algorithmic hyperparameters identical to the defaults.
Unless otherwise stated, all compression baselines were executed in a one-shot manner (i.e., without additional post-hoc fine-tuning), and we used a uniform expert retention rate of 62.5\% for Qwen3 (reducing experts from 128 to 80 per MoE layer) to ensure fair comparison.

\paragraph{REAP (Expert Pruning).}
We used the authors’ official implementation of REAP and kept all pruning-related hyperparameters at their default values.
For calibration, we adopted the same calibration data recipe used in the REAP paper: a 50/50 mixture of \texttt{allenai/c4} and \texttt{theblackcat102/evol-codealpaca-v1}.
All pruning runs for REAP were performed on two NVIDIA A100 40GB GPUs.

\paragraph{CFES (Expert Pruning).}
We used the authors' official implementation of CFES, and followed the default settings except for unavoidable runtime/distributed configurations. For calibration, we adopted the same calibration dataset composition used in the CFES paper: \texttt{rstar\_coder}, \texttt{openr1\_math220}, and \texttt{c4} (using the official sampling/preprocessing routine). \textit{(These correspond to code, mathematical reasoning, and knowledge domains, respectively.)} 
All pruning runs were conducted on two NVIDIA A100 40GB GPUs. The main hyperparameters are as follows:
\begin{itemize}
    \item \texttt{batch\_size=32}
    \item \texttt{num\_routed\_expert=80}
    \item \texttt{feature\_form=segment\_mean}
    \item \texttt{metric\_name=l2}
    \item \texttt{max\_length=4096}
    \item \texttt{prune\_method=c2f}
\end{itemize}

\paragraph{MoBE (Expert Editing).}
We used the authors' official implementation of MoBE and adopted the same optimization setup as the default configuration, except for the following explicitly specified hyperparameters.
We set the number of basis matrices to $m{=}8$ (i.e., \texttt{Basis(Num\_B)=8}) and compressed the model using a single NVIDIA A100 40GB GPU.
The MoBE factorization was trained with:
\begin{itemize}
    \item \texttt{--num\_epochs 10000}
    \item \texttt{--batch\_size 32}
    \item \texttt{--num\_batches 4}
    \item \texttt{--learning\_rate 0.07}
    \item \texttt{--activation "silu"}
    \item \texttt{--truncation 768}
\end{itemize}
All remaining options (including preprocessing and layer-wise handling) followed the official defaults.

\paragraph{HC-SMoE (Expert Merging).}
We used the authors' official implementation of HC-SMoE and kept all hyperparameters at their default values, except for the following settings explicitly chosen to match our experimental protocol.
All merging runs were performed on eight NVIDIA A100 40GB GPUs.
We did not use dominant experts (\texttt{DOMINANT="no"}), computed expert similarity based on expert outputs (\texttt{SIM\_BASE="expert-output"}), and used \texttt{zipit} as the default merge method (\texttt{MERGE\_METHOD="zipit"}).
We ran HC-SMoE in normal mode (\texttt{MODE="normal"}) with the following configuration:
\begin{itemize}
    \item \texttt{Number\_SENTENCES=4}, \texttt{max\_block\_size=2048}, \texttt{TRAIN\_Batch\_Size=2}
    \item \texttt{START\_LAYER=0}, \texttt{GROUP\_LIMIT=4}
    \item \texttt{CLUSTER\_METHOD="hierarchical"}, 
    \item \texttt{LINKAGE\_METHOD="average"}
    \item \texttt{STOP\_METRIC="silhouette"}
    \item \texttt{INGREDIENT="act"}
\end{itemize}
All other implementation details (including calibration-data sampling and merging pipeline internals) followed the official defaults.

\paragraph{TD-MoE (Expert Editing; Tensor Decomposition).}
We used the official TD-MoE implementation and followed its default configuration unless explicitly stated.
For Qwen3-30B-A3B-Instruct-2507, we used \texttt{c4} as the calibration dataset and applied a \texttt{global} clustering strategy (\texttt{CLUSTER\_TYPE="global"}).
We set the target compression ratio to \texttt{RATIO=0.4} and compressed the following MoE layers:
\begin{itemize}
    \item \texttt{LAYERS\_TO\_COMPRESS = \{4,5,7,8,9,10,11,13,14,15,18,33,34,35,36,37,38,39\}} 
\end{itemize}
The TD-MoE decomposition was run with:
\begin{itemize}
    \item \texttt{--whitening\_nsamples 256}
    \item \texttt{--cluster\_type "global"}
    \item \texttt{--model\_seq\_len 2048}
    \item \texttt{--whiten\_type "output"}
    \item \texttt{--layers\_to\_compress} (as above)
    \item \texttt{--ratio 0.4}
    \item \texttt{--decomposition\_method "svd"}
\end{itemize}
All remaining settings (including data sampling and runtime/distributed configurations) followed the official defaults, with only minimal adjustments required by our server environment.

\paragraph{M-SMoE (Expert Merging).}
We used the official M-SMoE implementation and kept the default settings unless explicitly stated.
We used \texttt{c4} as the calibration dataset and performed merging on eight NVIDIA A100 40GB GPUs.
We computed expert similarity based on router logits and used a small calibration subset:
\begin{itemize}
    \item \texttt{--similarity\_base="router-logits"}
    \item \texttt{--subset\_ratio=0.01}
\end{itemize}
The remaining key configuration follows:
\begin{itemize}
    \item \texttt{block\_size = 512}
    \item \texttt{batch\_size = 1}
    \item \texttt{num\_fewshot = 5}
\end{itemize}
All other hyperparameters and the end-to-end merging pipeline followed the official defaults, except for unavoidable runtime/distributed adjustments to match our hardware constraints.

\subsection{Common protocol in Mixtral.}
For all experiments on the Mixtral backbone, we followed the default configurations provided by each baseline’s official code release as closely as possible. We only introduced minimal, unavoidable changes to accommodate our server constraints (e.g., number of GPUs and runtime/distributed settings), while keeping the algorithmic hyperparameters unchanged unless explicitly specified below. 
Unless otherwise stated, all compression baselines were executed in a one-shot manner (i.e., without additional post-hoc fine-tuning). 
For pruning/merging-style baselines on Mixtral, we used a uniform expert retention rate of 62.5\% (reducing experts from 8 to 5 per MoE layer) for fair comparison.

\paragraph{REAP (Expert Pruning).}
We used the authors’ official implementation of REAP and kept all pruning-related hyperparameters at their default values.
For calibration, we adopted the same calibration data recipe used in the REAP paper: a 50/50 mixture of \texttt{allenai/c4} and \texttt{theblackcat102/evol-codealpaca-v1}.
All pruning runs for REAP were conducted on two NVIDIA A100 80GB GPUs.

\paragraph{CFES (Expert Pruning).}
We used the authors’ official implementation of CFES and followed the default settings except for unavoidable runtime/distributed configurations.
For calibration, we adopted the same calibration dataset composition used in the CFES paper: \texttt{rstar\_coder}, \texttt{openr1\_math220}, and \texttt{c4} (using the official sampling/preprocessing routine).
All pruning runs for CFES were conducted on two NVIDIA A100 80GB GPUs. The main hyperparameters are:
\begin{itemize}
    \item \texttt{batch\_size=32}
    \item \texttt{num\_routed\_expert=5} 
    \item \texttt{feature\_form=segment\_mean}
    \item \texttt{metric\_name=l2}
    \item \texttt{max\_length=4096}
    \item \texttt{prune\_method=c2f}
\end{itemize}

\paragraph{MoBE (Expert Editing).}
For MoBE, we followed the official implementation and kept the default settings except for the basis count and the listed training hyperparameters.
We set the number of bases to $\texttt{Num\_B}=2$ and compressed the model on one NVIDIA A100 80GB GPU.
The MoBE factorization was trained with:
\begin{itemize}
    \item \texttt{--num\_epochs 10000}
    \item \texttt{--batch\_size 32}
    \item \texttt{--num\_batches 4}
    \item \texttt{--learning\_rate 0.07}
    \item \texttt{--activation "silu"}
    \item \texttt{--truncation 1672}
\end{itemize}
All remaining options (including preprocessing and layer-wise handling) followed the official defaults.

\paragraph{HC-SMoE (Expert Merging).}
We used the authors' official implementation of HC-SMoE and kept all hyperparameters at their default values, except for the following settings explicitly chosen to match our experimental protocol.
All merging runs were performed on eight NVIDIA A100 80GB GPUs.
\begin{itemize}
    \item \texttt{DOMINANT="no"} (no dominant expert)
    \item \texttt{SIM\_BASE="expert-output"} (expert-output-based similarity)
    \item \texttt{MERGE\_METHOD="zipit"} (default merge method)
    \item \texttt{MODE="normal"}
    \item \texttt{Number\_SENTENCES=32}, 
    \item \texttt{max\_block\_size=2048}, 
    \item \texttt{TRAIN\_Batch\_Size=2}
    \item \texttt{START\_LAYER=0}, 
    \item \texttt{GROUP\_LIMIT=4}
    \item \texttt{CLUSTER\_METHOD="hierarchical"}, 
    \item \texttt{LINKAGE\_METHOD="average"}
    \item \texttt{STOP\_METRIC="silhouette"}
    \item \texttt{INGREDIENT="act"}
\end{itemize}
All other implementation details (including calibration-data sampling and merging pipeline internals) followed the official defaults.

\paragraph{TD-MoE (Tensor Decomposition).}
We used the authors’ official TD-MoE implementation with \texttt{c4} as the calibration dataset and \texttt{global} clustering (\texttt{CLUSTER\_TYPE="global"}). 
We set the target compression ratio to \texttt{RATIO=0.4} and compressed the following Mixtral MoE layers:
\begin{itemize}
    \item \texttt{LAYERS\_TO\_COMPRESS = (3, 5, 6, 7, 9, 10, 12, 22, 23, 24, 25, 26)}.
\end{itemize}
The decomposition was run with the following key arguments:
\begin{itemize}
    \item \texttt{MODEL\_PATH="mistralai/Mixtral-8x7B-Instruct-v0.1"}
    \item \texttt{DATASET="c4"}, 
    \item \texttt{--cluster\_type "global"}, 
    \item \texttt{--ratio 0.4}
    \item \texttt{--whitening\_nsamples 256}
    \item \texttt{--model\_seq\_len 2048}
    \item \texttt{--whiten\_type "output"}
    \item \texttt{--layers\_to\_compress} (as above)
    \item \texttt{--decomposition\_method "svd"}
\end{itemize}
All remaining options followed the official defaults, with only minimal runtime/distributed adjustments required by our server environment.

\paragraph{M-SMoE (Expert Merging).}
We used the authors’ official M-SMoE implementation and preserved the default settings except for unavoidable runtime/distributed configurations.
Merging was performed on eight NVIDIA A100 80GB GPUs.
We computed expert similarity based on router logits and used a small calibration subset:
\begin{itemize}
    \item \texttt{--similarity\_base="router-logits"}
    \item \texttt{--subset\_ratio=0.01}
\end{itemize}
The remaining key configuration is:
\begin{itemize}
    \item \texttt{block\_size=512}
    \item \texttt{batch\_size=1}
    \item \texttt{subset\_ratio=0.01}
    \item \texttt{num\_fewshot=5}
\end{itemize}
All other hyperparameters and merging pipeline details followed the authors’ defaults, except for unavoidable runtime/distributed adjustments to match our hardware constraints.

\subsection{Router Knowledge Distillation Hyperparameter Settings}

\noindent\textbf{Note.} The Router KD hyperparameters and the calibration dataset (\texttt{c4}) \cite{c4} are identical across \emph{all} experimental cases. (See Table~\ref{tab:router_kd_hparams_shared})

\begin{table}[h]
\centering
\caption{\textbf{Router KD hyperparameters (shared across all cases).}
We used \emph{exactly the same} Router KD hyperparameters for \emph{every} experiment case and model variant, and consistently used \texttt{c4} as the calibration dataset; only the teacher--student pair (i.e., the backbone and the compressed baseline) was changed.}
\setlength{\tabcolsep}{7pt}
\begin{tabular}{l l}
\hline
\textbf{Parameter} & \textbf{Value} \\
\hline
Calibration dataset & \texttt{c4} \\
Epochs & 1 \\
Batch size & 2 \\
Gradient accumulation steps & 4 \\
Learning rate & $5\times 10^{-5}$ \\
Max sequence length & 512 \\
KD temperature ($T$) & 1.0 \\
Max calibration samples & 3000 \\
\hline
\end{tabular}
\vspace{2pt}
\label{tab:router_kd_hparams_shared}
\end{table}

\newpage

\section{Experiment Result Tables}
\label{appendix:exp_tables}

\input{tables/qwen3_table_router_375_1st}

\input{tables/qwen3_table_router_375_2nd}

\input{tables/mixtral_table_router_r5_1st}

\input{tables/mixtral_table_router_r5_2nd}

\input{tables/qwen3_table_router_25_3rd}

\end{document}

%% file: tables/qwen3_table_router_375_1st.tex
\begin{table*}[h]
\centering
\caption{Performance comparison of compression methods with Router KD on \textit{Qwen3-30B-A3B-Instruct-2507}. This table compares the performance of representative methods from each category—Expert Pruning (REAP), Expert Editing (MoBE), and Expert Merging (HC-SMoE)—against their Router KD calibrated versions (denoted with -$R$). The results demonstrate that Router KD consistently recovers performance across diverse benchmarks on the fine-grained \textit{Qwen3-30B-A3B-Instruct-2507} architecture.}
\label{tab:table1}

\small
\setlength{\tabcolsep}{6pt}
\renewcommand{\arraystretch}{1.15}

\begin{tabular}{l|l|
                S[table-format=1.4]|
                S[table-format=1.4]
                S[table-format=1.4]|
                S[table-format=1.4]
                S[table-format=1.4]|
                S[table-format=1.4]
                S[table-format=1.4]}
\toprule
& & {\textbf{Original}}
  & {\textbf{REAP}}
  & {\textbf{REAP-\textit{R}}}
  & {\textbf{MoBE}}
  & {\textbf{MoBE-\textit{R}}}
  & {\textbf{HC-SMoE}}
  & {\textbf{HC-SMoE-\textit{R}}} \\
\midrule
& Method        & Original & {Prune} & {Prune} & {Edit} & {Edit} & {Merge} & {Merge} \\
& \# Total Params & {30.53B} & {19.66B}   & {19.66B}& {19.66B}& {19.66B}& {19.66B}& {19.66B} \\
\midrule

\multirow{3}{*}{General}    & BBH-Fewshot   & 0.3039 & \best{0.4866}    & 0.4810        & 0.0023       & \best{0.4715} & 0.3107 & \best{0.3173}  \\
                            & BBH-Zeroshot  & 0.4222 & 0.4446    & \best{0.4468}        & 0.4016       & \best{0.4128} & 0.3665 & \best{0.3726}  \\
                            & CoQA          & 0.4283 & \best{0.4107}    & 0.4015        & 0.2553       & \best{0.2820} & \best{0.3688} & 0.3642  \\
\midrule

                      & GSM8k          & 0.8628 & 0.8802    & \best{0.8832}        & 0.8006       & \best{0.8575} & 0.7339 & \best{0.7498}  \\
\multirow{3}{*}{Math} & GSM8k Platinum & 0.8875 & 0.8999    & \best{0.9090}        & 0.8354       & \best{0.8776} & 0.7585 & \best{0.7750}  \\
                      & MATH           & 0.2712 & 0.2736    & \best{0.2786}        & 0.2438       & \best{0.4608} & 0.2518 & \best{0.2578}  \\
                      & AIME 1983-2024 & 0.3912 & \best{0.3483}    & 0.3451        & 0.2069       & \best{0.2347} & 0.1854 & \best{0.1919}  \\
                      & AIME 2025      & 0.2333 & \best{0.2667}    & 0.2333            & 0.1333       & 0.1333        & 0.0667 & \best{0.1667}  \\
\midrule

\multirow{2}{*}{Coding} & MBPP               & 0.6780 & \best{0.6220}    & 0.6180        & \best{0.5860}       & 0.5640 & \best{0.6340} & 0.6240  \\
                        & HumanEval-instruct & 0.9390 & 0.9146    & \best{0.9207}        & 0.8049       & \best{0.8171} & 0.8841 & \best{0.8963}  \\
\midrule

\multirow{3}{*}{CoT-Fewshot}    & BBH            & 0.0488 & 0.0644    & \best{0.0660}        & 0.0103       & \best{0.4181} & 0.0364 & \best{0.0367}  \\
                                & GSM8k          & 0.8241 & \best{0.8757}    & 0.8749        & 0.7847       & \best{0.8355} & 0.7506 & \best{0.7650}  \\
                                & GSM8k Platinum & 0.8594 & 0.9065    & \best{0.9074}        & 0.8147       & \best{0.8718} & 0.7783 & \best{0.7883}  \\
\midrule

\multirow{3}{*}{CoT-Zeroshot}   & BBH            & 0.3863 & 0.3850    & \best{0.3853}        & 0.4065       & \best{0.4170} & 0.3130 & \best{0.3288}  \\
                                & GSM8k          & 0.5671 & \best{0.6149}    & 0.6080        & \best{0.6277}       & 0.6217 & 0.5868 & \best{0.5951}  \\
                                & GSM8k Platinum & 0.6005 & \best{0.6460}    & 0.6394        & \best{0.6634}       & 0.6543 & 0.6179 & \best{0.6352}  \\
\midrule

                              & ARC-challenge   & 0.4787 & \best{0.4215}    & 0.4181        & \best{0.4462}       & 0.4369 & 0.3788 & \best{0.3899}  \\
                              & ARC-easy        & 0.7256 & 0.6936    & \best{0.6957}        & 0.6991       & \best{0.7138} & 0.6549 & \best{0.6599}  \\
                              & HellaSwag       & 0.4153 & 0.4198    & \best{0.4245}        & 0.4429       & \best{0.4522} & 0.3684 & \best{0.3778}  \\
\multirow{3}{*}{Multi-Choice} & MedMCQA         & 0.5420 & 0.4100    & \best{0.4109}        & 0.4109       & \best{0.4282} & \best{0.3371} & 0.3359  \\
                              & MedQA           & 0.4556 & 0.4218    & \best{0.4266}        & 0.3174       & \best{0.3244} & 0.2883 & \best{0.2914}  \\
                              & OpenbookQA      & 0.3320 & 0.2960    & \best{0.3020}        & 0.2740       & \best{0.2860} & 0.2260 & \best{0.2360}  \\
                              & PIQA            & 0.7258 & 0.7203    & \best{0.7247}        & 0.7356       & \best{0.7383} & \best{0.6534} & 0.6496  \\
                              & WinoGrande      & 0.5683 & 0.5683    & \best{0.5841}        & 0.5706       & \best{0.5848} & 0.5391 & \best{0.5493}  \\
                              & MMLU            & 0.7112 & 0.6483    & \best{0.6507}        & \best{0.6369}       & 0.6347 & 0.4313 & \best{0.4517}  \\  

\bottomrule
\end{tabular}
\end{table*}

%% file: tables/qwen3_table_router_375_2nd.tex
\begin{table*}[t]
\centering
\caption{Performance comparison of alternative compression baselines with Router KD on \textit{Qwen3-30B-A3B-Instruct-2507}. We evaluate additional baselines for each category: CFES (Pruning), TD-MoE (Editing), and M-SMoE (Merging). Consistent with Table~\ref{tab:table1}, applying Router KD (denoted with -$R$) yields performance improvements across most tasks, reinforcing the generalizability of our proposed calibration strategy.}
\label{tab:main_compare}

\small
\setlength{\tabcolsep}{6pt}
\renewcommand{\arraystretch}{1.15}

\begin{tabular}{l|l|
                S[table-format=1.4]|
                S[table-format=1.4]
                S[table-format=1.4]|
                S[table-format=1.4]
                S[table-format=1.4]|
                S[table-format=1.4]
                S[table-format=1.4]}
\toprule
& & {\textbf{Original}}
  & {\textbf{CFES}}
  & {\textbf{CFES-\textit{R}}}
  & {\textbf{TD-MoE}}
  & {\textbf{TD-MoE-\textit{R}}}
  & {\textbf{M-SMoE}}
  & {\textbf{M-SMoE-\textit{R}}} \\
\midrule
& Method        & Original & {Prune} & {Prune} & {Edit} & {Edit} & {Merge} & {Merge} \\
& \# Total Params & {30.53B} & {19.66B}   & {19.66B}& {19.66B}& {19.66B}& {19.66B}& {19.66B} \\
\midrule

\multirow{3}{*}{General}    & BBH-Fewshot   & 0.3039 & \best{0.3416} & 0.3285  & \best{0.3446} & 0.3442 & 0.3758 & \best{0.3791}  \\
                            & BBH-Zeroshot  & 0.4222 & 0.3843 & \best{0.3992}  & 0.4105 & \best{0.4147} & 0.4124 & \best{0.4136}  \\
                            & CoQA          & 0.4283 & \best{0.2458} & 0.2443  & \best{0.3095} & 0.2990 & 0.3097 & \best{0.3252}  \\
\midrule

                      & GSM8k          & 0.8628 & 0.6209 & \best{0.6649}  & 0.8052 & \best{0.8127} & 0.5330 & \best{0.5527}  \\
\multirow{3}{*}{Math} & GSM8k Platinum & 0.8875 & 0.6443 & \best{0.7047}  & 0.8478 & \best{0.8536} & 0.5633 & \best{0.5806}  \\
                      & MATH           & 0.2712 & 0.2786 & \best{0.2950}  & 0.2748 & \best{0.2794} & 0.0330 & \best{0.0336}  \\
                      & AIME 1983-2024 & 0.3912 & 0.0986 & \best{0.1565}  & 0.1800 & \best{0.1844} & 0.0000 & 0.0000  \\
                      & AIME 2025      & 0.2333 & 0.0667 & \best{0.1000}  & 0.1333 & 0.1333 & 0.0000 & 0.0000  \\
\midrule

\multirow{2}{*}{Coding} & MBPP               & 0.6780 & 0.1040 & \best{0.1860}  & 0.5620 & \best{0.5740} & 0.0000 & 0.0000  \\
                        & HumanEval-instruct & 0.9390 & 0.0183 & \best{0.0854}  & \best{0.8171} & 0.8110 & 0.0000 & 0.0000  \\
\midrule

\multirow{3}{*}{CoT-Fewshot}    & BBH            & 0.0488 & \best{0.3281} & 0.3231  & \best{0.0264} & 0.0256 & \best{0.1921} & 0.1599  \\
                                & GSM8k          & 0.8241 & 0.6133 & \best{0.6641}  & 0.8256 & \best{0.8317} & 0.5625 & \best{0.5732}  \\
                                & GSM8k Platinum & 0.8594 & 0.6336 & \best{0.6940}  & 0.8553 & \best{0.8619} & 0.5864 & \best{0.5972}  \\
\midrule

\multirow{3}{*}{CoT-Zeroshot}   & BBH            & 0.3863 & 0.3571 & \best{0.3734}  & 0.3634 & \best{0.3655} & 0.3070 & \best{0.3090}  \\
                                & GSM8k          & 0.5671 & 0.3927 & \best{0.4466}  & \best{0.5709} & 0.5679 & 0.1736 & \best{0.1979}  \\
                                & GSM8k Platinum & 0.6005 & 0.4152 & \best{0.4682}  & 0.5955 & \best{0.5997} & 0.1844 & \best{0.2126}  \\
\midrule

                              & ARC-challenge   & 0.4787 & \best{0.3183} & 0.3166  & 0.4352 & \best{0.4394} & 0.4147 & \best{0.4232}  \\
                              & ARC-easy        & 0.7256 & 0.5109 & \best{0.5253}  & 0.6848 & \best{0.6890} & 0.6827 & \best{0.6928}  \\
                              & HellaSwag       & 0.4153 & 0.4534 & \best{0.4753}  & \best{0.4655} & 0.4639 & 0.4456 & \best{0.4532}  \\
\multirow{3}{*}{Multi-Choice} & MedMCQA         & 0.5420 & 0.3015 & \best{0.3330}  & 0.3832 & \best{0.3870} & 0.4898 & \best{0.4994}  \\
                              & MedQA           & 0.4556 & 0.3048 & \best{0.3229}  & 0.2891 & \best{0.2899} & 0.4643 & \best{0.4941}  \\
                              & OpenbookQA      & 0.3320 & 0.2060 & \best{0.2300}  & \best{0.2820} & 0.2780 & \best{0.3260} & 0.3200  \\
                              & PIQA            & 0.7258 & 0.7220 & \best{0.7383}  & 0.7497 & \best{0.7519} & 0.7252 & \best{0.7307}  \\
                              & WinoGrande      & 0.5683 & 0.5872 & \best{0.6212}  & \best{0.5801} & 0.5691 & \best{0.5896} & 0.5872  \\
                              & MMLU            & 0.7112 & 0.5486 & \best{0.5614}  & 0.4803 & \best{0.4920} & 0.6696 & \best{0.6716}  \\

\bottomrule
\end{tabular}
\end{table*}

%% file: tables/mixtral_table_router_r5_1st.tex
\begin{table*}[t]
\centering
\caption{Performance comparison of compression methods with Router KD on \textit{Mixtral-8x7B-Instruct-v0.1}. This table presents the evaluation results using the coarse-grained Mixtral architecture as the backbone. We compare REAP, MoBE, and HC-SMoE with their Router KD counterparts. As discussed in Section~\ref{sec:inefficient}, the performance gains from Router KD are relatively marginal compared to \textit{Qwen3-30B-A3B-Instruct-2507}, due to the simpler routing decision boundaries of the coarse-grained architecture.}
\label{tab:table3}

\small
\setlength{\tabcolsep}{6pt}
\renewcommand{\arraystretch}{1.15}

\begin{tabular}{l|l|
                S[table-format=1.4]|
                S[table-format=1.4]
                S[table-format=1.4]|
                S[table-format=1.4]
                S[table-format=1.4]|
                S[table-format=1.4]
                S[table-format=1.4]}
\toprule
& & {\textbf{Original}}
  & {\textbf{REAP}}
  & {\textbf{REAP-\textit{R}}}
  & {\textbf{MoBE}}
  & {\textbf{MoBE-\textit{R}}}
  & {\textbf{HC-SMoE}}
  & {\textbf{HC-SMoE-\textit{R}}} \\
\midrule
& Method        & Original & {Prune} & {Prune} & {Edit} & {Edit} & {Merge} & {Merge} \\
& \# Total Params & {46.70B} & {29.79B}   & {29.79B}& {29.79B}& {29.79B}& {29.79B}& {29.79B} \\
\midrule

\multirow{3}{*}{General}    & BBH-Fewshot   & 0.0000  & 0.0000        & 0.0000 & 0.0000 & 0.0000 & 0.0000 & 0.0000  \\
                            & BBH-Zeroshot  & 0.4661  & \best{0.2224} & 0.2133 & \best{0.2992} & 0.2953 & \best{0.4038} & 0.4030  \\
                            & CoQA          & 0.1698  & 0.3145 & \best{0.3400} & \best{0.3083} & 0.2805 & 0.1342 & \best{0.1468}  \\
\midrule

                      & GSM8k          & 0.6247 & 0.0311 & \best{0.0334}        & 0.0243       & \best{0.0318} & 0.3859 & \best{0.3920}  \\
\multirow{3}{*}{Math} & GSM8k Platinum & 0.6592 & 0.0306 & \best{0.0339}        & \best{0.0331}       & 0.0314 & 0.3921 & \best{0.4078}  \\
                      & MATH           & 0.2798 & \best{0.0346} & 0.0308        & 0.0428       & \best{0.0432} & 0.1026 & \best{0.1170}  \\
                      & AIME 1983-2024 & 0.0021 & 0.0000 & 0.0000               & 0.0000       & 0.0000 & 0.0000 & 0.0000  \\
                      & AIME 2025      & 0.0000 & 0.0000 & 0.0000               & 0.0000       & 0.0000 & 0.0000 & 0.0000  \\
\midrule

\multirow{2}{*}{Coding} & MBPP                   & 0.0000 & 0.0000 & 0.0000               & 0.0000       & 0.0000 & 0.0000 & 0.0000  \\
                        & HumanEval-instruct     & 0.5427 & 0.0000 & \best{0.0122}        & \best{0.0122}      & 0.0000 & 0.2683 & \best{0.2927}  \\
\midrule

\multirow{3}{*}{CoT-Fewshot}    & BBH            & 0.6481 & 0.2657 & \best{0.2780}        & 0.1605       & \best{0.1754} & 0.5228 & \best{0.5356}  \\
                                & GSM8k          & 0.6892 & 0.0364 & 0.0364               & \best{0.0826}       & 0.0781 & 0.3700 & \best{0.3882}  \\
                                & GSM8k Platinum & 0.7146 & 0.0372 & \best{0.0521}        & \best{0.0902}       & 0.0794 & 0.3772 & \best{0.3978}  \\
\midrule

\multirow{3}{*}{CoT-Zeroshot}    & BBH           & 0.4909 & 0.1193 & \best{0.1539}       & 0.2500       & \best{0.2520} & \best{0.4360} & 0.4319  \\
                                & GSM8k          & 0.6050 & 0.0227 &\best{ 0.0288}       & \best{0.1130}       & 0.0910 & 0.3692 & \best{0.3783}  \\
                                & GSM8k Platinum & 0.6154 & 0.0265 & 0.0265              & \best{0.1117}       & 0.0935 & 0.3879 & 0.3879  \\
\midrule

                              & ARC-challenge   & 0.5026 & 0.2816 & \best{0.3020}        & 0.4155       & \best{0.4249} & 0.4983 & \best{0.5068}  \\
                              & ARC-easy        & 0.7189 & 0.4566 & \best{0.4575}        & \best{0.6797}       & 0.6734 & \best{0.7677} & 0.7668  \\
                              & HellaSwag       & 0.6364 & 0.3981 & \best{0.4008}        & \best{0.5043}       & 0.4870 & 0.6065 & \best{0.6092}  \\
\multirow{3}{*}{Multi-Choice} & MedMCQA         & 0.5377 & 0.2678 & \best{0.2728}        & \best{0.3753}       & 0.3610 & \best{0.4052} & 0.4028  \\
                              & MedQA           & 0.5727 & 0.2584 & \best{0.2773}        & \best{0.3943}       & 0.3661 & 0.4273 & \best{0.4297}  \\
                              & OpenbookQA      & 0.3440 & 0.1940 & \best{0.2240}        & 0.3240       & \best{0.3300} & 0.3480 & \best{0.3560}  \\
                              & PIQA            & 0.7448 & 0.5827 & \best{0.5892}        & \best{0.7388}       & 0.7296 & \best{0.7807} & 0.7791  \\
                              & WinoGrande      & 0.6156 & \best{0.5627} & 0.4996        & 0.6093       & \best{0.6290} & \best{0.6875} & 0.6811  \\
                              & MMLU            & 0.6765 & 0.2829 & \best{0.2874}        & 0.4515       & \best{0.4623} & 0.5753 & \best{0.5758}  \\    

\bottomrule
\end{tabular}
\end{table*}

%% file: tables/mixtral_table_router_r5_2nd.tex
\begin{table*}[t]
\centering
\caption{Performance comparison of alternative compression baselines with Router KD on \textit{Mixtral-8x7B-Instruct-v0.1}. Evaluation of CFES, TD-MoE, and M-SMoE on the Mixtral backbone. Similar to Table~\ref{tab:table3}, the results show the impact of Router KD on a coarse-grained MoE model. While some recovery is observed, it highlights the structural limitations of router calibration in models with fewer experts.}
\label{tab:main_compare}

\small
\setlength{\tabcolsep}{6pt}
\renewcommand{\arraystretch}{1.15}

\begin{tabular}{l|l|
                S[table-format=1.4]|
                S[table-format=1.4]
                S[table-format=1.4]|
                S[table-format=1.4]
                S[table-format=1.4]|
                S[table-format=1.4]
                S[table-format=1.4]}
\toprule
& & {\textbf{Original}}
  & {\textbf{CFES}}
  & {\textbf{CFES-\textit{R}}}
  & {\textbf{TD-MoE}}
  & {\textbf{TD-MoE-\textit{R}}}
  & {\textbf{M-SMoE}}
  & {\textbf{M-SMoE-\textit{R}}} \\
\midrule
& Method        & Original & {Prune} & {Prune} & {Edit} & {Edit} & {Merge} & {Merge} \\
& \# Total Params & {46.70B} & {29.79B}   & {29.79B}& {29.79B}& {29.79B}& {29.79B}& {29.79B} \\
\midrule

\multirow{3}{*}{General}    & BBH-Fewshot   & 0.0000  & 0.0000           & 0.0000        & 0.0000 & 0.0000  & 0.0000 & 0.0000  \\
                            & BBH-Zeroshot  & 0.4643  & \best{0.2049}    & 0.1955        & \best{0.3986} & 0.3961 & 0.2264 & \best{0.3110}  \\
                            & CoQA          & 0.1668  & 0.2888           & \best{0.2938} & 0.1878 & \best{0.1895} & \best{0.1832} & 0.0130  \\
\midrule

                      & GSM8k          & 0.6361 & \best{0.0197} & 0.0182        & 0.3753       & \best{0.3844}        & 0.0546 & \best{0.1380}  \\
\multirow{3}{*}{Math} & GSM8k Platinum & 0.6592 & 0.0149 & \best{0.0199}        & 0.4028              & \best{0.4185} & 0.0571 & \best{0.1266}  \\
                      & MATH           & 0.2834 & 0.0236 & \best{0.0260}        & 0.1442              & \best{0.1500} & 0.0306 & \best{0.0400}  \\
                      & AIME 1983-2024 & 0.0021 & 0.0000 & 0.0000        & 0.0000              & 0.0000        & 0.0000 & 0.0000  \\
                      & AIME 2025      & 0.0000 & 0.0000 & 0.0000        & 0.0000              & 0.0000        & 0.0000 & 0.0000  \\
\midrule

\multirow{2}{*}{Coding} & MBPP               & 0.0000 & 0.0000 & 0.0000        & 0.0000       & 0.0000        & 0.0000 & 0.0000  \\
                        & HumanEval-instruct & 0.5427 & 0.0000 & 0.0000        & 0.3476       & 0.3476 & 0.0976 &  \best{0.1402}  \\
\midrule

\multirow{3}{*}{CoT-Fewshot}    & BBH            & 0.6481 & \best{0.2646} & 0.2542        & 0.4970              & \best{0.5004} & 0.3130 & \best{0.3706}  \\
                                & GSM8k          & 0.6892 & 0.0273 & \best{0.0303}        & \best{0.3601}       & 0.3472        & 0.0614 & \best{0.1274}  \\
                                & GSM8k Platinum & 0.7146 & \best{0.0265} & 0.0207        & \best{0.3672}       & 0.3631        & 0.0620 & \best{0.1373}  \\
\midrule

\multirow{3}{*}{CoT-Zeroshot}   & BBH            & 0.4909 & 0.1146 & \best{0.1181}        & \best{0.4072}       & 0.4006 & 0.2321 & \best{0.2786}  \\
                                & GSM8k          & 0.6050 & 0.0174 & \best{0.0182}        & 0.3859              & 0.3859 & 0.0743 & \best{0.1175}  \\
                                & GSM8k Platinum & 0.6154 & 0.0141 & \best{0.0149}        & 0.4003              & \best{0.4111} & 0.0835 & \best{0.1141}  \\
\midrule

                              & ARC-challenge   & 0.5026 & 0.2543 & \best{0.2611}        & \best{0.4915}       & 0.4753 & 0.3942 & \best{0.4394}  \\
                              & ARC-easy        & 0.7189 & \best{0.4019} & 0.3851        & \best{0.7218}       & 0.7210 & 0.6650 & \best{0.6801}  \\
                              & HellaSwag       & 0.6364 & \best{0.3795} & 0.3782        & 0.6137       & \best{0.6138} & 0.4954 & \best{0.5368}  \\
\multirow{3}{*}{Multi-Choice} & MedMCQA         & 0.5377 & \best{0.2768} & 0.2704        & \best{0.4260}       & 0.4236 & 0.3325 & \best{0.3352}  \\
                              & MedQA           & 0.5727 & 0.2482 & \best{0.2608}        & \best{0.4690}       & 0.4635 & 0.3056 & \best{0.3307}  \\
                              & OpenbookQA      & 0.3440 & \best{0.1840} & 0.1780        & 0.3220       & \best{0.3380} & 0.2580 & \best{0.3360}  \\
                              & PIQA            & 0.7448 & \best{0.5696} & 0.5664        & \best{0.7622}       & 0.7541 & 0.7247 & \best{0.7563} \\
                              & WinoGrande      & 0.6156 & \best{0.5359} & 0.5241        & \best{0.6393}       & 0.6361 & 0.6488 & \best{0.6598}  \\
                              & MMLU            & 0.6765 & 0.2481 & \best{0.2512}       & 0.5649       & \best{0.5670} & 0.3825 & \best{0.4284}  \\    

\bottomrule
\end{tabular}
\end{table*}

%% file: tables/qwen3_table_router_25_3rd.tex
\begin{table*}[h]
\centering
\caption{Robustness analysis of Router KD under milder compression (75\% Parameter Retention) on \textit{Qwen3-30B-A3B-Instruct-2507}. To verify that the efficacy of Router KD is not limited to a specific compression ratio, we evaluated performance while retaining 75\% of the expert parameters ($\approx$ 23.28B total parameters). The results confirm that Router KD consistently mitigates performance degradation even under this alternative compression setting.}
\label{tab:additional}

\small
\setlength{\tabcolsep}{6pt}
\renewcommand{\arraystretch}{1.15}

\begin{tabular}{l|l|
                S[table-format=1.4]|
                S[table-format=1.4]
                S[table-format=1.4]|
                S[table-format=1.4]
                S[table-format=1.4]|
                S[table-format=1.4]
                S[table-format=1.4]}
\toprule
& & {\textbf{Original}}
  & {\textbf{CFES}}
  & {\textbf{CFES-\textit{R}}}
  & {\textbf{MoBE}}
  & {\textbf{MoBE-\textit{R}}}
  & {\textbf{HC-SMoE}}
  & {\textbf{HC-SMoE-\textit{R}}} \\
\midrule
& Method        & Original & {Prune} & {Prune} & {Edit} & {Edit} & {Merge} & {Merge} \\
& \# Total Params & {30.53B} & {23.28B}   & {23.28B}& {23.28B}& {23.28B}& {23.28B}& {23.28B} \\
\midrule

\multirow{3}{*}{General}    & BBH-Fewshot   & 0.3039 & 0.2268    & \best{0.2273}        & 0.0026       & \best{0.3575} & 0.3000 & \best{0.3018}  \\
                            & BBH-Zeroshot  & 0.4222 & 0.3764    & \best{0.4162}        & 0.4165       & \best{0.4170} & 0.3755 & \best{0.3812}  \\
                            & CoQA          & 0.4283 & 0.1462    & \best{0.1810}        & \best{0.3565}       & 0.3412 & \best{0.3955} & 0.3953  \\
\midrule

                      & GSM8k          & 0.8628 & 0.4117    & \best{0.6475}        & \best{0.8658}       & 0.8582 & 0.8393 & \best{0.8415}  \\
\multirow{3}{*}{Math} & GSM8k Platinum & 0.8875 & 0.4342    & \best{0.6782}        & 0.8892       & \best{0.8900} & 0.8635 & \best{0.8668}  \\
                      & MATH           & 0.2712 & 0.2274    & \best{0.2824}        & 0.2396       & \best{0.2940} & 0.2696 & \best{0.2726}  \\
                      & AIME 1983-2024 & 0.3912 & 0.0086    & \best{0.0911}        & \best{0.3891}       & 0.3805 & 0.3708 & 0.3708  \\
                      & AIME 2025      & 0.2333 & 0.0000    & \best{0.0333}        & 0.2333       & 0.2333        & 0.2667 & 0.2667  \\
\midrule

\multirow{2}{*}{Coding} & MBPP               & 0.6780 & 0.1140    & \best{0.4720}        & 0.4440       & \best{0.6480} & \best{0.6500} & 0.6440  \\
                        & HumanEval-instruct & 0.9390 & 0.1341    & \best{0.7012}        & \best{0.9085}       & 0.8963 & 0.9146 & 0.9146  \\
\midrule

\multirow{3}{*}{CoT-Fewshot}    & BBH            & 0.0488 & 0.3098    & \best{0.3797}        & \best{0.0361}       & 0.0359 & 0.0412 & \best{0.0429}  \\
                                & GSM8k          & 0.8241 & 0.4579    & \best{0.6300}        & \best{0.8461}       & 0.8378 & 0.8340 & \best{0.8408}  \\
                                & GSM8k Platinum & 0.8594 & 0.4806    & \best{0.6609}        & \best{0.8768}       & 0.8685 & 0.8644 & \best{0.8726}  \\
\midrule

\multirow{3}{*}{CoT-Zeroshot}   & BBH            & 0.3863 & 0.3213    & \best{0.3422}        & 0.3738       & \best{0.3774} & \best{0.3672} & 0.3671  \\
                                & GSM8k          & 0.5671 & 0.2631    & \best{0.4776}        & \best{0.5588}       & 0.5519 & 0.6353 & \best{0.6482}  \\
                                & GSM8k Platinum & 0.6005 & 0.2796    & \best{0.5136}        & \best{0.5947}       & 0.5889 & 0.6741 & \best{0.6791}  \\
\midrule

                              & ARC-challenge   & 0.4787 & 0.3609    & \best{0.3831}        & 0.4753       & \best{0.4821} & 0.4480 & \best{0.4505}  \\
                              & ARC-easy        & 0.7256 & 0.5467    & \best{0.6237}        & \best{0.7273}       & 0.7269 & 0.7235 & \best{0.7294}  \\
                              & HellaSwag       & 0.4153 & 0.4655    & \best{0.4982}        & \best{0.4588}       & 0.4584 & 0.4212 & \best{0.4245}  \\
\multirow{3}{*}{Multi-Choice} & MedMCQA         & 0.5420 & 0.3622    & \best{0.3918}        & 0.5078       & \best{0.5104} & 0.3662 & \best{0.3739}  \\
                              & MedQA           & 0.4556 & 0.2969    & \best{0.3771}        & 0.3841       & \best{0.3881} & 0.3480 & \best{0.3535}  \\
                              & OpenbookQA      & 0.3320 & 0.2380    & \best{0.2780}        & \best{0.3040}       & 0.3000 & 0.2840 & \best{0.3020} \\
                              & PIQA            & 0.7258 & 0.7307    & \best{0.7508}        & \best{0.7486}       & 0.7476 & \best{0.7057} & 0.7018  \\
                              & WinoGrande      & 0.5683 & 0.5991    & \best{0.6030}        & 0.5841       & \best{0.5896} & 0.5746 & \best{0.5833}  \\
                              & MMLU            & 0.7112 & 0.5472    & \best{0.6288}        & \best{0.6958}       & 0.6954 & 0.5691 & \best{0.5766}  \\  

\bottomrule
\end{tabular}
\end{table*}